\documentclass{article}

\PassOptionsToPackage{numbers,sort&compress}{natbib}
\usepackage[preprint]{neurips_2026}

\usepackage[utf8]{inputenc}
\usepackage[T1]{fontenc}
\usepackage{amsmath,amssymb,amsfonts}
\usepackage{booktabs}
\usepackage{graphicx}
\usepackage{multirow}
\usepackage{xcolor}
\usepackage{xspace}
\usepackage{enumitem}
\usepackage{float}
\usepackage[section]{placeins}
\usepackage{doi}

\newcommand{\yhat}{\hat{y}}
\newcommand{\sigmarho}{\sigma(\rho)}
\newcommand{\lambdaxi}{\lambda_\xi}
\newcommand{\xistar}{\xi^*}
\newcommand{\Sbar}{\bar{S}}
\newcommand{\sigmapds}{\hat{\sigma}_{\text{PDS}}}

\title{Escaping the Agreement Trap:\\
Defensibility Signals for Evaluating Rule-Governed AI}

\author{
  Michael O'Herlihy\thanks{Equal contribution.} \\
  Reddit, Inc. \\
  \texttt{michael.oherlihy@reddit.com} \\
  \And
  Rosa Catal\`{a}\footnotemark[1] \\
  Reddit, Inc. \\
  \texttt{rosa.catala@reddit.com} \\
}

\begin{document}

\maketitle

\begin{abstract}
Content moderation systems are typically evaluated by measuring agreement with human labels.
In rule-governed environments this assumption fails: multiple decisions may be logically
consistent with the governing policy, and agreement metrics penalize valid decisions while
mischaracterizing ambiguity as error---a failure mode we term the \emph{Agreement Trap}.
We formalize evaluation as policy-grounded correctness and introduce the Defensibility
Index~(DI) and Ambiguity Index~(AI). To estimate reasoning stability without additional
audit passes, we introduce the Probabilistic Defensibility Signal~(PDS), derived from
audit-model token logprobs. We harness LLM reasoning traces as a governance signal rather
than a classification output by deploying the audit model not to decide whether content
violates policy, but to verify whether a proposed decision is logically derivable from the
governing rule hierarchy.
We validate the framework on 193{,}000+ Reddit moderation decisions across multiple
communities and evaluation cohorts, finding a \textbf{33--46.6 percentage-point gap} between agreement-based and
policy-grounded metrics, with \textbf{79.8--80.6\% of the model's false negatives corresponding to
policy-grounded decisions} rather than true errors. We further show that measured ambiguity is driven by rule
specificity: auditing 37{,}286 identical decisions under three tiers of the same community
rules reduces AI by 10.8~pp while DI remains stable. Repeated-sampling analysis attributes
PDS variance primarily to governance ambiguity rather than decoding noise. A Governance Gate built on these signals achieves \textbf{78.6\% automation
coverage with 64.9\% risk reduction}. Together, these results show that evaluation in
rule-governed environments should shift from agreement with historical labels to
reasoning-grounded validity under explicit rules.
\end{abstract}

%=============================================================================
\section{Introduction}
\label{sec:intro}
%=============================================================================

Automated content moderation systems operating under explicit governance rules face an
evaluation problem that agreement-based metrics cannot solve. When a platform's rules
permit multiple distinct decisions for a given piece of content, as they frequently do in
grey areas, human moderators resolve the ambiguity using context, precedent, and
interpretive norms in concert with the written
policy~\citep{aroyo2015truth,davani2022dealing,pavlick2019inherent,uma2021learning}. A
model that learns from these labels learns to reproduce the interpretive pathway, not the
rule structure. Measured by $F_1$ it appears to perform well; measured against the governing
rules it may be systematically less aligned than a model with lower agreement scores.

We term this the \emph{Agreement Trap}. It arises whenever correctness is set-valued: when
the set of policy-defensible outcomes contains more than one element, agreement penalizes
valid decisions and conflates three distinct failure modes---model error, moderator
divergence, and policy ambiguity---into a single undifferentiated
signal~\citep{plank2022problem,uma2021learning}. Each requires a different intervention;
none is distinguishable from the others under~$F_1$.

Legal theory has long recognized that rule systems contain a \emph{penumbra} of interpretive
ambiguity where multiple outcomes are defensible under the same rule
structure~\citep{hart1961concept}. Rules exist at varying levels of
specificity---from headline titles to detailed guidelines with examples and
exceptions---and the interpretive latitude available to moderators depends on which layer
is visible, a dynamic we term \emph{Normative Underspecification}. This observation is
consistent with prior work showing that moderation practice is often tacit and
contextual~\citep{chandrasekharan2018hidden,gillespie2018custodians,gorwa2020algorithmic}.

Policy ambiguity is not a deficiency but a deliberate design feature of principle-based
rules. It preserves human moderator flexibility and judgment in novel or unexpected
situations without requiring constant policy rewrites. Policies that are overly specific
can ironically create loopholes; Reddit's principle-based approach keeps rules elastic
enough to cover new contexts. The Defensibility Framework and Ambiguity Index do not treat
this elasticity as error. Instead, they quantify it as a measurable governance property so
that automation can be deployed safely and transparently precisely where defensibility is
high, while preserving the human judgment that the rules were designed to enable.

We make three contributions.
\textbf{First}, we formalize evaluation as policy-grounded correctness and introduce DI and
AI as metrics that separate policy-defensible from indefensible decisions.
\textbf{Second}, we introduce PDS, a logprob-based stability signal extracted at zero
additional cost from the audit model's reasoning trace.
\textbf{Third}, we operationalize DI and AI in a Governance Gate that achieves principled
automation thresholds with empirically validated coverage-risk tradeoffs. We validate on
Reddit's production moderation infrastructure and present adversarial analysis of the
framework's failure modes.

We validate the framework on 193{,}000+ offline model decisions spanning multiple
Reddit communities and cohorts, audited against both platform-wide rules~($R_G$) and
community-specific rules~($R_C$); no decision in this study was used for live
enforcement. We find a large gap between agreement-based and policy-grounded evaluation,
with most model false negatives corresponding to policy-grounded decisions rather than true
errors. We further show that PDS provides a calibrated reasoning-stability signal, that its
variance is driven primarily by governance ambiguity rather than sampling noise, and that a
Governance Gate built on these signals achieves substantial automation coverage with
meaningful risk reduction. Finally, adversarial analysis shows that the framework's escape
surface aligns with the same policy-ambiguity regions identified by the Ambiguity Index.

%=============================================================================
\section{Related Work}
\label{sec:related}
%=============================================================================

\paragraph{Disagreement as signal.}
\citet{aroyo2015truth} argue that annotator disagreement often reflects genuine task
ambiguity rather than error; \citet{davani2022dealing} show this in subjective annotation
settings such as hate speech; \citet{uma2021learning} survey the broader literature on
learning from disagreement; \citet{plank2022problem} argues that label variation should be
treated as a first-class modeling and evaluation concern; and \citet{pavlick2019inherent}
show that disagreement can be inherent to the task itself.

\paragraph{LLM-as-judge.}
LLM-based evaluation typically raises calibration and self-enhancement concerns when models
assess outputs against implicit or subjective
standards~\citep{zheng2023judging,liu2026examining}. Our approach differs structurally from
these paradigms. The audit model is not asked to evaluate whether an output is \emph{good}
or preferable; it evaluates whether a proposed decision~$\yhat$ is logically derivable from
an explicit rule system~$(R, P)$. The governing rules are treated as the external source of
truth, and the audit model functions as a constrained reasoning engine over this rule set.
The task therefore has a formal answer: given $(C, R, P, \yhat)$, does a valid derivation
exist? This distinction is critical---the audit model is not substituting for human judgment
or preference, but is operationalizing rule-consistency under an explicit policy. By ensuring
that the model commits to a rule citation before assigning a defensibility level, errors are
characterized as failures of derivation rather than disagreements with subjective intent.
Circularity is addressed through four independent triangulation channels
(Section~\ref{sec:circularity}). This shifts the evaluation target from preference judgment
to reasoning-grounded validity under an explicit rule structure.

\paragraph{Logprob uncertainty.}
\citet{kadavath2022language} show that language models can assess their own uncertainty via
logprobs; \citet{kuhn2023semantic} extend this to semantic uncertainty. PDS differs by
extracting uncertainty at reasoning-critical token positions rather than at the output label,
and by conditioning on an explicit governance structure. \citet{lanham2023measuring} and
\citet{turpin2023language} motivate extracting uncertainty over chain-of-thought tokens
specifically.

\paragraph{Calibration and selective prediction.}
We follow standard Expected Calibration Error (ECE) methodology~\citep{guo2017calibration,naeini2015well}
and adapt selective prediction~\citep{geifman2017selective} to governance-aware deployment.

%=============================================================================
\section{The Defensibility Framework}
\label{sec:framework}
%=============================================================================

\subsection{Formalization}
\label{sec:formalization}

Let $C$ be the content under review, $R = (R_G, R_C)$ the applicable rule hierarchy
(platform-wide $R_G$, community-specific $R_C$), $P$ the community precedent corpus, and
$\yhat \in \{\textsc{remove}, \textsc{approve}\}$ a proposed decision. This formalization
treats moderation evaluation as derivability under an explicit rule structure, in the spirit
of rule-based reasoning traditions in legal theory and AI \&
law~\citep{hart1961concept}. A decision $\yhat$ is \emph{defensible} under $(C, R, P)$ if a
valid logical derivation $D$ from $(C, R, P)$ to $\yhat$ exists---a chain of inference
steps grounded in the explicit rule text, requiring no premises absent from $(C, R, P)$.
Defensibility levels:

\begin{itemize}[leftmargin=2em]
  \item \textbf{L1---Robustly Defensible}: an explicit rule directly and unambiguously
    authorizes~$\yhat$.
  \item \textbf{L2---Plausibly Defensible}: rules are genuinely ambiguous about this
    scenario but could reasonably support~$\yhat$.
  \item \textbf{L3---Indefensible}: no explicit rule authorizes~$\yhat$, or the content
    complies with the literal requirements of the cited rule, or the reasoning invokes
    concepts absent from the rule set.
\end{itemize}

\begin{equation}
  \text{Defensibility Index (DI)} =
    \frac{|\{i : \text{level}(i) \in \{L_1, L_2\}\}|}{N}.
  \quad\text{Target: DI} \geq 0.90.
  \label{eq:di}
\end{equation}

\begin{equation}
  \text{Ambiguity Index (AI)} =
    \frac{|\{i : \text{inverse\_check}(i) = \texttt{Yes}\}|}{N},
  \quad\text{Target: AI} \leq 0.15.
  \label{eq:ai}
\end{equation}

AI decomposes into $\mathrm{AI_G}$ (platform-level underspecification in $R_G$) and
$\mathrm{AI_C}$ (community-level underspecification in $R_C$). These components indicate
different interventions: $\mathrm{AI_G}$ suggests platform policy revision; $\mathrm{AI_C}$
suggests community precedent development. The Normative Underspecification experiment
(Section~\ref{sec:shadowing}) provides an empirical estimate of $\mathrm{AI_C}$ by
varying rule specificity within a single community.

\subsection{The Agreement Trap}
\label{sec:agreement_trap}

The Agreement Trap arises when $\text{AI} > 0$: human moderators must resolve ambiguity
using context in concert with the written rules, producing labels that reflect the resolution
pathway rather than the rule structure~\citep{plank2022problem,uma2021learning}. A model trained on
these labels learns the pathway, not the derivation. The empirical signature is false
negatives that are L1 or L2---agreement-based evaluation penalizing policy adherence. If
80\% of the model's false negatives are defensible, the evaluation is measuring interpretive
alignment rather than rule-grounded validity.

\subsection{Audit Implementation}
\label{sec:audit_impl}

The framework is operationalized through a structured audit. Audit model $M_a$ receives
$(C, R, P, \yhat)$ and generates a JSON reasoning trace with fields in this order:
$\texttt{logic\_chain} \to \texttt{policy\_citation}~(\kappa) \to
\texttt{precedent\_weight}~(\omega \in \{\text{High}, \text{Medium}, \text{Low}\}) \to
\texttt{inverse\_check}~(\iota \in \{\text{Yes}, \text{No}\}) \to
\texttt{defensibility\_level}~(\xi \in \{1,2,3\})$.
This ordering is a deliberate design choice: committing to a rule citation before assigning
defensibility level ensures that classification is grounded in an explicit governance
reference. $M_a$ is not generating a moderation decision---it is auditing whether $\yhat$ is
derivable from $(R, P)$.

\begin{figure}[H]
  \centering
  \includegraphics[width=\textwidth]{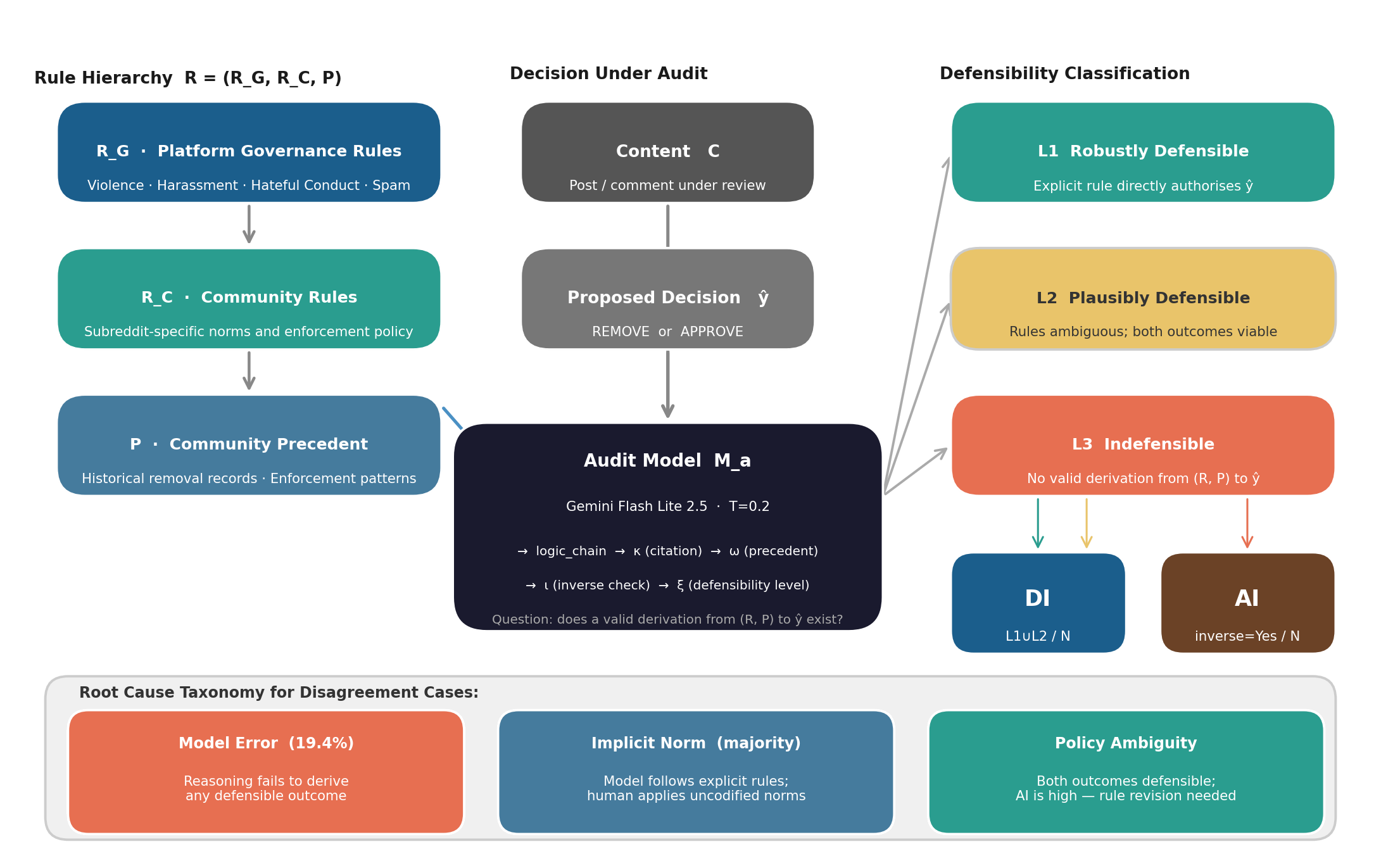}
  \caption{The Defensibility Framework. Rule hierarchy $R = (R_G, R_C, P)$ flows into audit
  model $M_a$ alongside content $C$ and proposed decision~$\yhat$. $M_a$ generates trace
  $A = (\kappa, \omega, \iota, \xi)$ and classifies $\yhat$ as L1 (Robustly Defensible),
  L2 (Plausibly Defensible), or L3 (Indefensible). DI aggregates the $L_1 \cup L_2$
  fraction; AI aggregates cases where the inverse check fires. Root cause taxonomy
  decomposes disagreement into Model Error (19.4\%), Implicit Norm Enforcement, and Policy
  Ambiguity.}
  \label{fig:framework}
\end{figure}

The Governance Gate restricts automated enforcement to decision cohorts satisfying
$\text{DI} \geq 0.90$, $\text{AI} \leq 0.15$ (minimum 25 decisions). The threshold pair
was selected as the knee of the coverage-risk tradeoff from scenario sensitivity analysis
(Table~\ref{tab:gate}).

\subsection{Operational Deployment}
\label{sec:deployment}

The Defensibility Framework can be integrated into existing moderation systems with minimal
architectural changes. The following pipeline describes a reference integration; the
framework has not been deployed in production at the time of writing.

\begin{enumerate}[leftmargin=2em]
  \item \textbf{Audit pass.} For each decision~$\yhat$ produced by a moderation
    model~$M_c$, run the audit model~$M_a$ on $(C, R, P, \yhat)$ to produce a JSON
    reasoning trace.
  \item \textbf{Metric computation.} Aggregate defensibility outcomes to compute the
    Defensibility Index ($L_1 \cup L_2$ rate) and Ambiguity Index (inverse-check rate)
    over decision cohorts.
  \item \textbf{Ambiguity diagnosis.} Identify high-AI regions to distinguish
    platform-level ($\mathrm{AI_G}$) or community-level ($\mathrm{AI_C}$) policy
    underspecification from true model error.
  \item \textbf{Governance Gate.} Restrict automated enforcement to decision cohorts
    satisfying $\text{DI} \geq 0.90$ and $\text{AI} \leq 0.15$, selecting operating
    points along the coverage--risk frontier (Section~\ref{sec:gate}).
\end{enumerate}

\noindent
This pipeline converts moderation evaluation from agreement-based scoring to
policy-grounded control, enabling the safe expansion of automation while preserving
auditability.

%=============================================================================
\section{Probabilistic Defensibility Signal}
\label{sec:pds}
%=============================================================================

\subsection{Motivation and Architecture}
\label{sec:pds_motivation}

Standard output confidence fails as a stability signal: it remains uniformly high
(0.95--0.99) even during extreme classification reversals. This occurs because confidence is
measured at the final output token, after the model has committed to a reasoning
path~\citep{kadavath2022language,kuhn2023semantic}. PDS instead extracts uncertainty at
token positions that precede the defensibility verdict, motivated by evidence that
chain-of-thought outputs may not faithfully reflect the underlying computation and can
behave like post-hoc
rationalizations~\citep{lanham2023measuring,turpin2023language}.

The architecture is two-model: $M_c$ produces $\yhat$; $M_a$ audits whether $\yhat$ is
derivable from $(R, P)$, with $\yhat$ as a fixed input. The \emph{Audit Independence
Assumption}~(AIA) holds when $M_c$ and $M_a$ are sufficiently independent that $M_a$ has no
systematic prior toward justifying $M_c$'s outputs. The \emph{Same-Backbone
Condition}~(SBC) is violated when they share a backbone, inflating PDS for labels consistent
with $M_c$'s prior. Calibration corrects operationally for SBC but the decomposition
interpretation (that PDS components proxy specific ambiguity types) is stronger under AIA.

\subsection{Three-Component Vector}
\label{sec:pds_components}

PDS extracts three scalars from $M_a$'s single forward pass:

\paragraph{$\lambdaxi$---Label log-confidence.}
$\lambdaxi = \log p_{\theta_a}(\xistar \mid C, R, P, \yhat)$ where
$\xistar = \arg\max_{l \in \{1,2,3\}} p_{\theta_a}(l \mid C, R, P, \yhat)$ is the MAP
defensibility level (argmax over the softmax distribution, not the sampled token; at
$T \geq 0.7$ the distinction matters). Generated last in the template, conditioning on the
full preceding trace. Nearly uninformative standalone but captures commitment failure.

\paragraph{$H[\kappa]$---Citation span entropy.}
Mean conditional entropy over the \texttt{policy\_citation} token span:
\begin{equation}
  H[\kappa] = \frac{1}{n}\sum_{i=1}^{n}
    H[\kappa_i \mid \kappa_{1:i-1}, C, R, P, \yhat],
  \label{eq:hkappa}
\end{equation}
where each term is the Shannon entropy of $M_a$'s next-token distribution at position $i$
in the citation span. Extracted by character-offset matching of the JSON field delimiter;
100\% detection rate across 56{,}883 cases (mean span 25 tokens). Leading indicator:
generated before~$\xi$. Proxy for~$\mathrm{AI_G}$.

\paragraph{$H[w]$---Precedent weight entropy.}
Shannon entropy of the softmax distribution over $\{\text{High}, \text{Medium},
\text{Low}\}$ at the \texttt{precedent\_weight} token position. Single-token proxy for
$H[\kappa]$, exploiting the template's structured categorical output. Both
operationalizations yield equivalent calibration behavior (see Section~\ref{sec:calibration}).

\paragraph{$\sigmarho$---Inverse-check log-odds.}
$\rho = \log p(\texttt{Yes} \mid \pi) - \log p(\texttt{No} \mid \pi)$ where
$\pi = [C; R; P; \yhat; \texttt{logic\_chain}; \kappa^*; \texttt{precedent\_weight}]$ is
the full token prefix at the \texttt{inverse\_check} position. Note: $\xistar$ is generated
after $\iota$ in the template and is not in~$\pi$. $\sigmarho$ is the primary discriminator
between L1 (committed rule correctly applied, low $\sigmarho$) and L3 (rule contradicted,
high $\sigmarho$). Intermediate in generation order: after $\kappa$ but before $\xi$. Proxy
for $\mathrm{AI_C}$ when $H[\kappa]$ is low.

\medskip\noindent
The full PDS vector:
\begin{equation}
  \text{PDS}_{M_a}(C, R, P, \yhat) = (\lambdaxi,\; -H[w],\; -\sigmarho),
  \label{eq:pds_vector}
\end{equation}
all oriented so higher values indicate greater stability. Scalar collapse:
\begin{equation}
  S = \exp\bigl[\alpha \cdot \lambdaxi + \beta \cdot (-H[w])
    + \gamma \cdot (-\sigmarho)\bigr],
  \quad \alpha + \beta + \gamma = 1,\; \alpha,\beta,\gamma > 0.
  \label{eq:scalar}
\end{equation}
$S$ is calibrated as $P(\text{Not Indefensible} \mid C, R, P, \yhat, M_a)$.

\subsection{Calibration}
\label{sec:calibration}

Weight vector $(\alpha^*, \beta^*, \gamma^*)$ is fit on the Balanced Sample ($N\!=\!19{,}899$)
by MLE against hard defensibility labels ($y_i = 1$ if L1/L2, $y_i = 0$ if L3), using
softmax reparameterization $(\alpha, \beta, \gamma) = \text{softmax}(u)$ and L-BFGS-B
optimization on unconstrained $u \in \mathbb{R}^3$, following standard post-hoc calibration
methodology~\citep{guo2017calibration,naeini2015well}.

\begin{table}[H]
  \centering
  \caption{Calibrated PDS weights, $N\!=\!19{,}899$ Balanced Sample. Results consistent
  across $H[w]$ and $H[\kappa]$.}
  \label{tab:weights}
  \small
  \begin{tabular}{@{}llllp{5.1cm}@{}}
    \toprule
    Component & Weight $H[w]$ & Weight $H[\kappa]$ & Role & Interpretation \\
    \midrule
    $\alpha$ ($\lambdaxi$) & 0.629 & 0.605 & Label confidence &
      Dominant signal (${\sim}63\%$) \\
    $\beta$ ($-H[\cdot]$) & 0.011 & 0.040 & Entropy &
      Near-zero---see below \\
    $\gamma$ ($-\sigmarho$) & 0.360 & 0.354 & Inverse check &
      Strong second signal (${\sim}36\%$) \\
    \midrule
    Loss (binary CE) & 0.313 & --- & --- &
      ECE $= 0.076$ (calibration set); held-out ECE $= 0.042$--$0.057$ across
      $T \in \{0.1, 0.3, 0.7, 1.0\}$ on Random Sample
      ($N\!=\!26{,}009$--$26{,}512$). See Section~\ref{sec:heldout_ece}. \\
    \bottomrule
  \end{tabular}
\end{table}

The near-zero weight on the entropy component ($\beta^* \approx 0$) is a structural result
of the framework's design, not a limitation of the signal. Under the binary calibration
target, which groups Robustly Defensible (L1) and Plausibly Defensible (L2) cases against
Indefensible (L3) cases, entropy primarily serves to distinguish L1 from L2. However,
because L2 cases are considered policy-grounded, entropy does not effectively separate them
from L3 errors. As a result, the optimal decision boundary collapses onto $\lambdaxi$ (label
log-confidence) and $\sigmarho$ (inverse-check log-odds), which directly capture the L2/L3
distinction. Consistency of $\beta^* \approx 0$ across two independent
operationalizations ($H[w]$ and $H[\kappa]$) confirms this reflects the binary target
structure, not a measurement artifact. The calibrated scalar is approximately
$S \approx \exp[0.63 \cdot \lambdaxi - 0.36 \cdot \sigmarho]$.

This reveals an important property of policy-grounded evaluation: ambiguity (L2) is not
treated as error. Therefore, features that characterize ambiguity do not contribute to
binary correctness prediction. A multi-class calibration that explicitly separates L1 and
L2---isolating rule-grounded from precedent-dependent defensibility---would be expected to
recover a non-zero role for entropy-based components, identifying a clear target for future
work.

\begin{figure}[H]
  \centering
  \includegraphics[width=\textwidth]{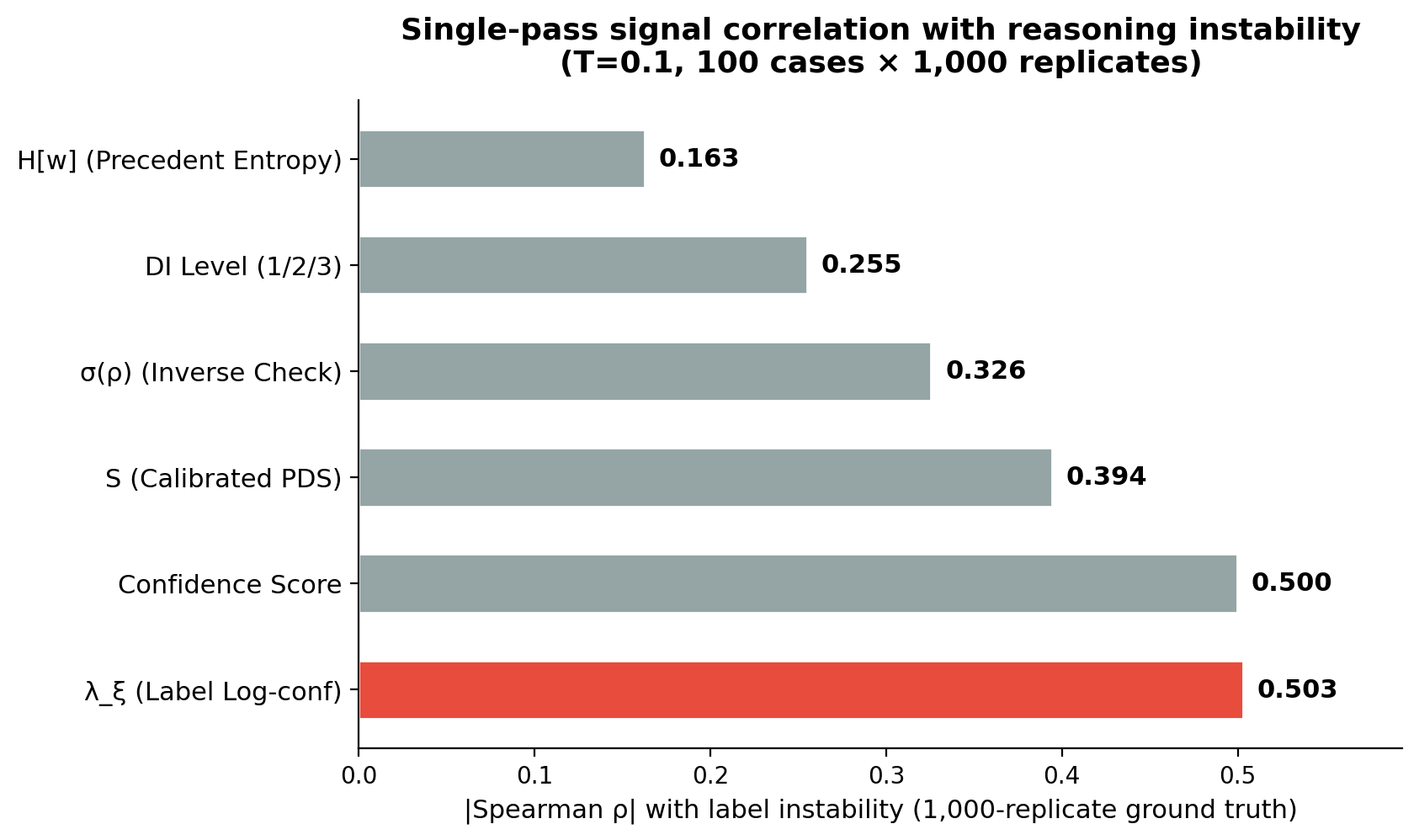}
  \caption{Single-pass signal correlation with reasoning instability ($T\!=\!0.1$, 100
  cases $\times$ 1{,}000 replicates). $\lambdaxi$ (label log-confidence) is the strongest
  single-pass predictor ($|\rho|\!=\!0.503$); $H[w]$ (precedent entropy) is weakest
  ($|\rho|\!=\!0.163$), explaining $\beta^* \approx 0$ under the binary calibration target.}
  \label{fig:signal_correlation}
\end{figure}

\begin{figure}[H]
  \centering
  \includegraphics[width=\textwidth]{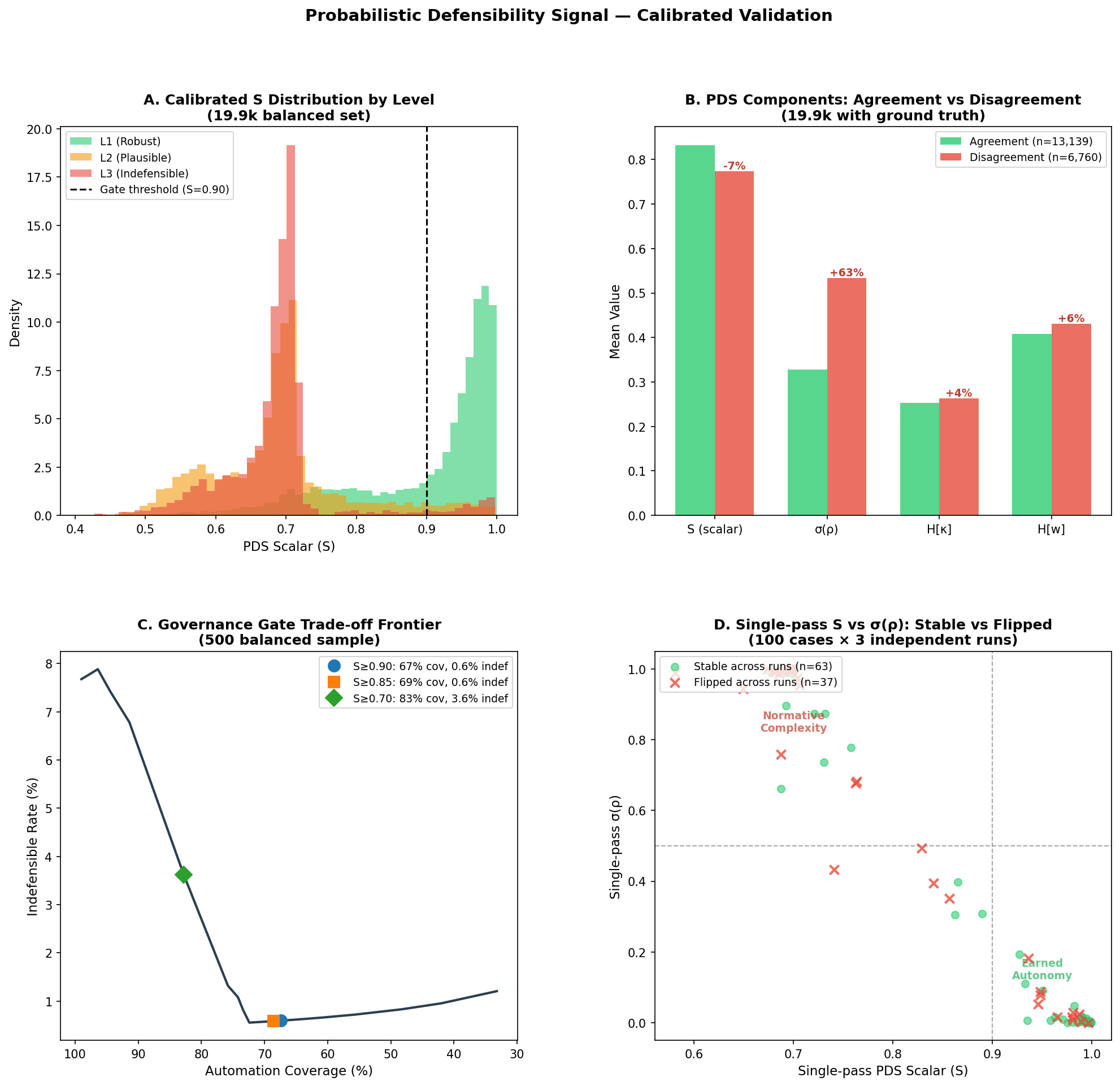}
  \caption{PDS calibration validation. (A)~Calibrated $S$ distribution by defensibility
  level. (B)~$\sigmarho$ is 63\% higher in disagreement vs agreement cases; $S$ drops 7\%.
  (C)~Governance Gate trade-off frontier: Standard threshold ($S \geq 0.90$) is the knee.
  (D)~Single-pass $S$ vs $\sigmarho$ with Earned Autonomy and Normative Complexity clusters
  labeled.}
  \label{fig:pds_validation}
\end{figure}

%=============================================================================
\section{Empirical Evaluation}
\label{sec:evaluation}
%=============================================================================

\subsection{Setup}
\label{sec:setup}

We evaluate on 193{,}000+ Reddit moderation decisions drawn from multiple
cohorts.\footnote{This work presents a research evaluation framework applied
retrospectively to historical moderation data. It is not integrated into Reddit's production
moderation systems and has not directly influenced individual moderation decisions.} Two
primary cohorts: Random Sample ($N\!=\!26{,}902$ valid audits across 398 communities, representative of production traffic)
and Balanced Sample ($N\!=\!19{,}899$ valid audits, overrepresenting contested cases with
ground-truth human annotations). Audit model: Gemini~2.5 Flash Lite at $T\!=\!0.2$.
Community-specific rules serve as $R_C$; Reddit's platform-wide content policy as $R_G$.

\subsection{Agreement--Defensibility Gap}
\label{sec:gap}

The 33--46~pp gap is the central result. The 79.8--80.6\% defensible false negative rate is
the Agreement Trap signature: agreement evaluation is penalizing policy adherence. The
\emph{Accurate but Indefensible} finding (5.8--6\%) reveals cases where human-model
agreement conceals shared policy failure---a pattern consistent with prior work showing that
moderation labels can encode systematic disagreement and
bias~\citep{sap2019risk,founta2018large}.

\begin{table}[H]
  \centering
  \caption{Agreement-based versus policy-grounded metrics.}
  \label{tab:agreement_gap}
  \small
  \begin{tabular}{@{}lrrl@{}}
    \toprule
    Metric & Random & Balanced & Interpretation \\
    \midrule
    $F_1$ (agreement-based) & 45.7\% & 54.3\% & Standard metric \\
    Defensibility Index (DI) & 92.3\% & 87.3\% & Policy alignment \\
    Gap (DI vs.\ $F_1$) & +46.6\,pp & +33\,pp & The Agreement Trap penalty \\
    Ambiguity Index (AI) & 18.3\% & 24.9\% & Structural ambiguity \\
    False negatives that are defensible & 79.8\% & 80.6\% & Agreement Trap evidence \\
    Accurate but indefensible & ${\sim}$6\% & 5.8\% & Agreement masks policy gap \\
    \bottomrule
  \end{tabular}
\end{table}

\begin{figure}[H]
  \centering
  \includegraphics[width=\textwidth]{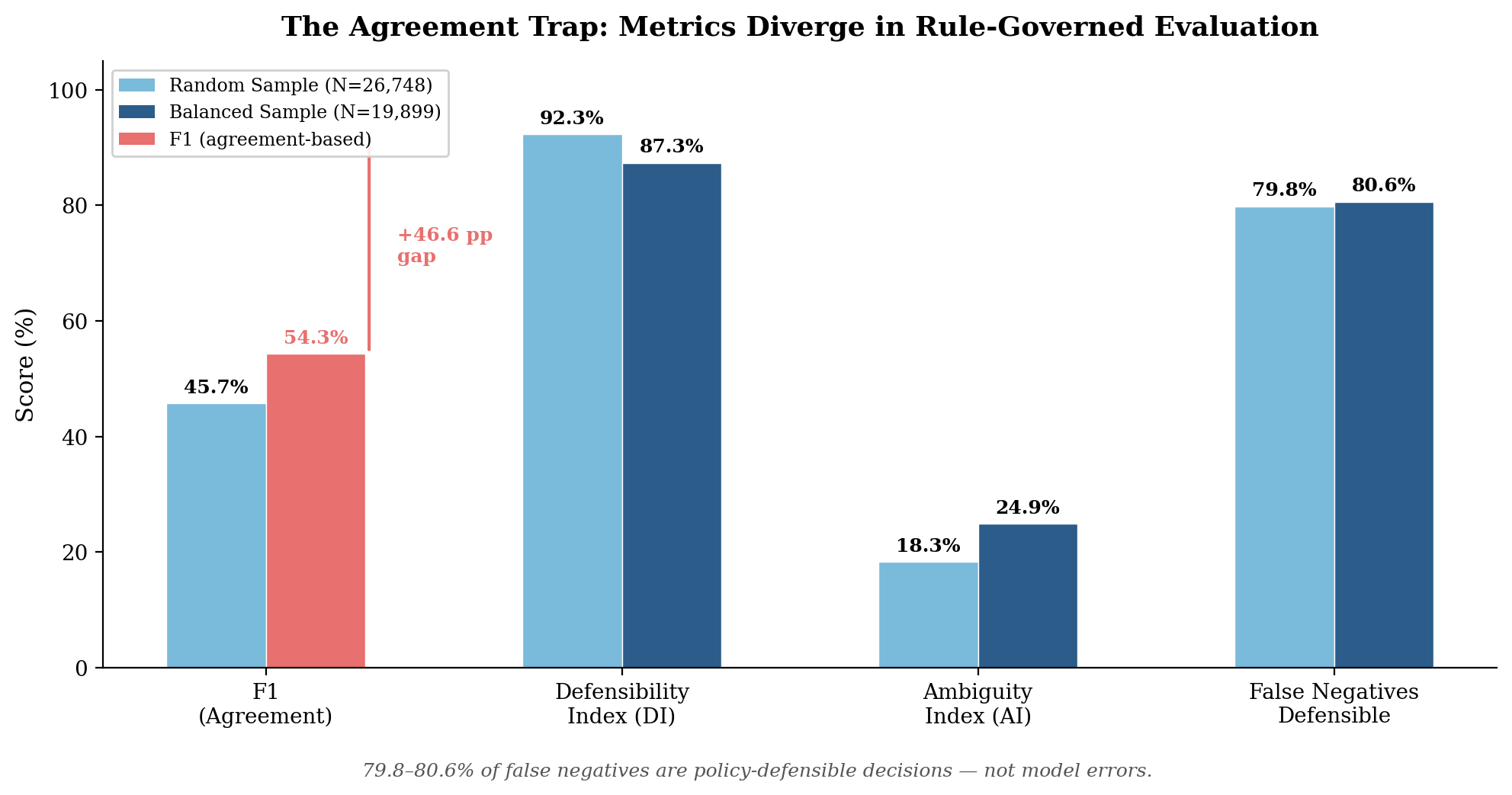}
  \caption{The Agreement--Defensibility Gap. $F_1$ (red) and Defensibility Index diverge by
  33--46 percentage points. 79.8--80.6\% of model false negatives are L1 or L2---policy-grounded
  decisions penalized by agreement-based evaluation.}
  \label{fig:agreement_gap}
\end{figure}

Root cause analysis of the 6{,}760 disagreement cases in the Balanced Sample: Model Error
(L3 classification) in 19.4\% (1{,}311 decisions); Policy-Grounded Disagreement in 80.6\%
(5{,}449 decisions) comprising \emph{Implicit Norm Enforcement} (model adheres to written
rules while human applies uncodified context) and \emph{Policy Ambiguity} (both outcomes
defensible, model and human chose different valid paths). Agreement-based evaluation counts
all 6{,}760 identically as model errors.

\subsection{Fleet-Level Diagnostics}
\label{sec:fleet}

Analysis of 270 communities (minimum 25 decisions) reveals three governance states:
\textbf{Earned Autonomy} ($N\!=\!165$, mean DI$\,=\,$96.8\%, AI$\,=\,$7.0\%),
\textbf{Policy Gaps} ($N\!=\!40$, DI$\,=\,$67.7\%, AI$\,=\,$36.0\%), and
\textbf{Normative Complexity} ($N\!=\!65$, DI$\,=\,$92.2\%, AI$\,=\,$23.0\%).
Normative Complexity communities are the critical case: the model reasons correctly but
operates in genuinely underspecified community-specific rule environments~($R_C$)---the AI
signal identifies where interpretive ambiguity is highest, not where model retraining is
needed. This refers
specifically to subreddit-level rules~($R_C$), not to platform-wide
policies~($R_G$).

\begin{figure}[H]
  \centering
  \includegraphics[width=\textwidth]{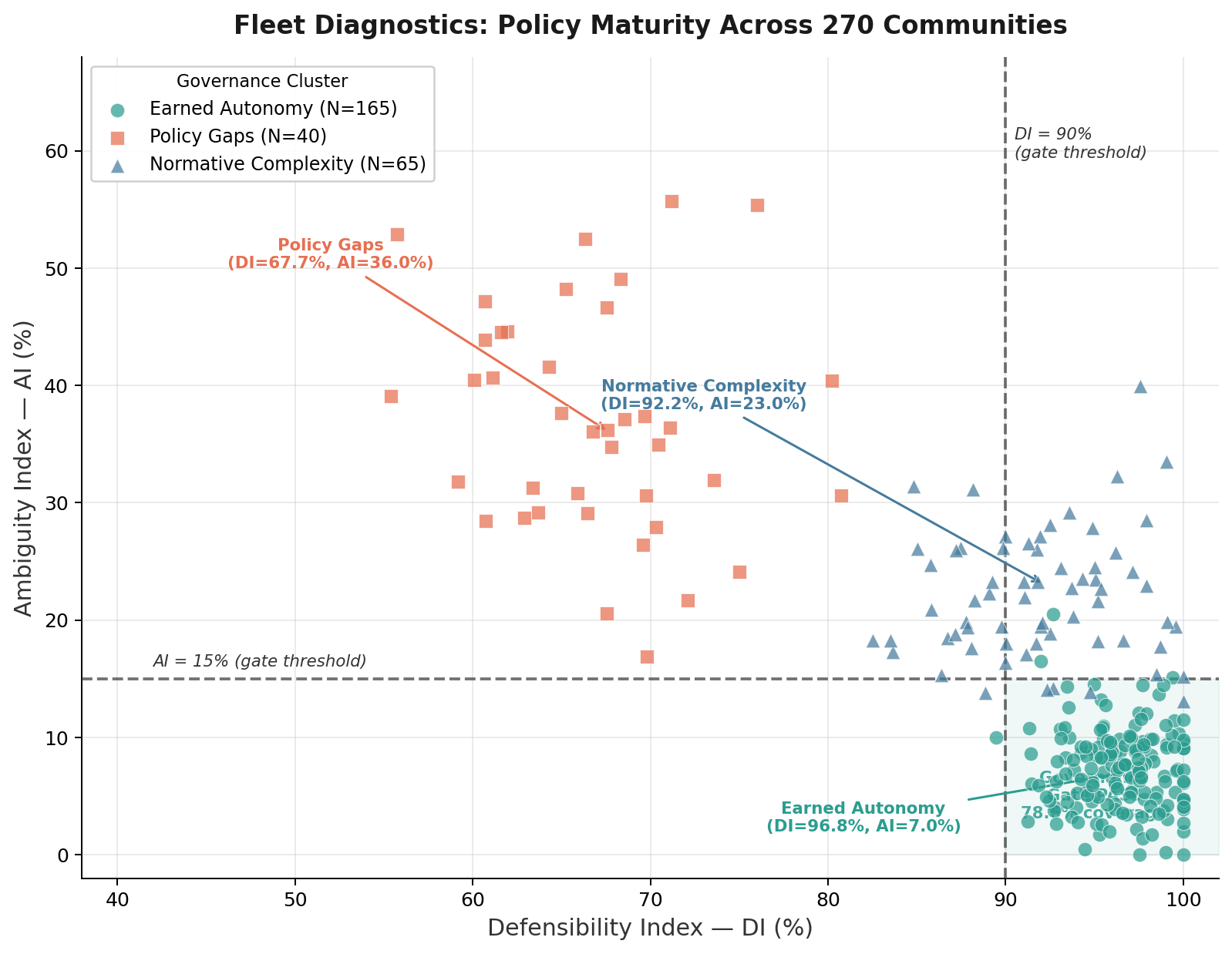}
  \caption{Fleet diagnostics across 270 communities. Earned Autonomy ($N\!=\!165$, circles)
  clusters in the Governance Gate pass zone ($\text{DI}\geq 90\%$, $\text{AI}\leq 15\%$); Policy Gaps
  ($N\!=\!40$, squares) shows reasoning failures; Normative Complexity ($N\!=\!65$,
  triangles) reasons correctly under underspecified rules. Dashed lines mark Standard Gate
  thresholds; shaded region is the 78.6\% coverage zone.}
  \label{fig:fleet}
\end{figure}

\subsection{PDS vs Standard Confidence}
\label{sec:pds_vs_conf}

$\sigmapds$ is 54\% higher in disagreement cases than agreement cases (mean
$\sigmapds\!=\!0.081$ vs $0.052$). Standard output confidence remains uniformly high
(0.95--0.99) in both regimes---it is uninformative about reasoning stability.

Human audit validation ($N\!=\!30$ cases, $K\!=\!121$ independent respondents): Spearman
$\rho\!=\!0.66$ between PDS and human defensibility ratings ($p < 10^{-4}$). DI-level
correlation $\rho\!=\!0.66$ ($p < 10^{-4}$), confirming strong alignment between model
and human assessments of policy-grounded validity.

$\sigmarho$ validates as a proxy for $\mathrm{AI_C}$ across two independent datasets. On the
Random Sample ($N\!\approx\!26.9$k), Spearman rank correlation between $\sigmarho$ and the
binary inverse check label is $\rho = 0.57$--$0.59$ ($p < 10^{-15}$) across
$T \in \{0.1, 0.3, 0.7, 1.0\}$, with mean $\sigmarho$ 6--7$\times$ higher for ambiguous
cases than unambiguous cases. On the Expert-Labeled Policy Set ($N\!=\!199$, single-policy
evaluation), $\rho = 0.19$ ($p = 0.007$). The weaker effect on this set reflects both
reduced sample size ($N\!=\!199$, only 31 ambiguous cases) and restricted policy scope
(single policy against a narrow rule set); the per-level structure is consistent across both
datasets: L1 shows low $\sigmarho$ and low AI; L2 shows high $\sigmarho$ and high AI; L3
shows the highest $\sigmarho$ and highest AI. Crucially, both datasets reach significance
independently and in the same direction.

The per-level structure confirms the paper's predictions exactly. L1 (Robustly Defensible):
AI $= 1.7$--$2.9\%$, mean $\sigmarho = 0.063$--$0.083$---a committed derivation forecloses
the opposite outcome. L2 (Plausibly Defensible): AI $= 58.8$--$61.4\%$, mean
$\sigmarho = 0.783$--$0.809$---genuine policy ambiguity where both outcomes are reachable.
L3 (Indefensible): AI $= 86.1$--$90.5\%$, mean $\sigmarho = 0.910$--$0.932$---the same
strong rule supports both actions, and the model's derivation contradicts it.

\begin{table}[H]
  \centering
  \caption{$\sigmarho$ as $\mathrm{AI_C}$ proxy---primary and secondary
  validation. $^{***}$\,$p < 10^{-15}$, $^{**}$\,$p < 0.01$. GS $=$ Expert-Labeled
  Policy Set (single-policy evaluation at $T\!=\!0.2$). $\sigmarho$ Yes/No $=$ mean
  $\sigmarho$ for cases where the inverse check fires vs does not. AI $=$ Ambiguity Index
  (fraction of cases where inverse check $=$ Yes).}
  \label{tab:sigma_rho}
  \small
  \begin{tabular}{@{}lrrcrrrr@{}}
    \toprule
    $T$ & $N$ & AI & Spearman $\rho$ & $p$ &
      $\sigmarho$ Yes & $\sigmarho$ No & Ratio \\
    \midrule
    $T\!=\!0.1$ & 26{,}935 & 22.2\% & $0.592^{***}$ & $<\!10^{-15}$ &
      0.894 & 0.127 & 7.04$\times$ \\
    $T\!=\!0.3$ & 26{,}858 & 22.5\% & $0.593^{***}$ & $<\!10^{-15}$ &
      0.900 & 0.129 & 6.95$\times$ \\
    $T\!=\!0.7$ & 26{,}611 & 22.4\% & $0.581^{***}$ & $<\!10^{-15}$ &
      0.907 & 0.142 & 6.38$\times$ \\
    $T\!=\!1.0$ & 26{,}519 & 22.8\% & $0.573^{***}$ & $<\!10^{-15}$ &
      0.905 & 0.151 & 5.97$\times$ \\
    \midrule
    GS & 199 & 15.6\% & $0.190^{**}$ & 0.007 &
      0.474 & 0.304 & 1.56$\times$ \\
    \bottomrule
  \end{tabular}
\end{table}

%=============================================================================
\section{Normative Underspecification}
\label{sec:shadowing}
%=============================================================================

We conduct a controlled rule-specificity experiment on $N\!=\!37{,}286$
decisions from r/AskReddit, one of Reddit's largest and longest-running communities.
r/AskReddit maintains an exceptionally well-developed governance structure: its
moderators have iteratively refined their rules over more than a decade, producing
documentation at three distinct layers of specificity---a rare example of mature,
community-driven policy evolution. This layered rule architecture makes r/AskReddit
an ideal natural experiment for isolating the effect of rule specificity on measured
ambiguity.

We audit each decision against three versions of the \emph{same} rules at increasing
levels of detail:
(i)~\textbf{Title Only}---rule headings alone (e.g., ``Rule~8: No
questions seeking professional advice''),
(ii)~\textbf{Sidebar}---headings plus brief descriptions and examples, and
(iii)~\textbf{Wiki}---full rule text with exceptions, edge cases, and worked examples.
All three runs use identical content and the same audit model; only the rule text varies.
This isolates rule specificity as the causal variable, ruling out the possibility that
DI/AI differences reflect differences in rule content rather than rule detail.

\begin{table}[H]
  \centering
  \caption{Normative Underspecification---effect of rule specificity on defensibility
  and ambiguity (r/AskReddit, $N\!=\!37{,}286$).}
  \label{tab:shadowing}
  \small
  \begin{tabular}{@{}lrrrr@{}}
    \toprule
    Rule Specificity & DI & AI & Indefensible & $\Delta$\,AI (from Title) \\
    \midrule
    Title Only          & 97.4\% & 18.2\% & 962  & ---         \\
    Sidebar (+\,Descriptions) & 98.0\% & 8.8\%  & 757  & $-9.4$\,pp  \\
    Wiki (+\,Examples/Exceptions) & 98.1\% & 7.4\%  & 722  & $-10.8$\,pp \\
    \bottomrule
  \end{tabular}
\end{table}

AI drops from 18.2\% to 8.8\% to 7.4\% as layers of rule detail are added. Each layer
reduces ambiguity, with the first layer (adding descriptions) doing the most work
($-9.4$\,pp). DI, by contrast, barely moves (97.4\% $\to$ 98.1\%)---the decisions are
defensible either way. The framework is not saying different things about the decisions;
it is saying the same thing with less uncertainty.

The effect is concentrated in removals: removal AI drops from 32.0\% (Title Only)
to 10.2\% (Sidebar and Wiki), while removal L1 rises from 48.5\% to 74.6\%.
Removals require affirmative justification---a specific rule must authorize the
action---so vague rules leave them stranded at L2. Approvals under prohibitive rules
are justified by the absence of a violation, making them inherently less sensitive to
rule specificity. The primary mechanism is L2$\to$L1 conversion: vague rules do not
create \emph{indefensible} decisions (L3 barely changes from 962 to 722); they create
\emph{ambiguous} ones.

The 10.8~pp AI reduction is our empirical estimate of community-level
underspecification ($\mathrm{AI_C}$): the fraction of decisions that are genuinely
contested under headline rules but become determinate when the full rule text is
available. This is consistent with prior work showing that moderation practice depends
heavily on tacit, contextual knowledge that is only partially discernable from
public-facing rule titles
alone~\citep{chandrasekharan2018hidden,gillespie2018custodians,gorwa2020algorithmic}.

%=============================================================================
\section{Stochastic Stability Analysis}
\label{sec:stability}
%=============================================================================

\subsection{Monte Carlo Estimator}
\label{sec:monte_carlo}

We estimate per-case reasoning stability by drawing $K\!=\!1{,}000$ independent samples from
$M_a$ at temperature $T$, computing $S$ for each replicate, and measuring
\begin{equation}
  \hat{\sigma}_{\text{PDS}} = \text{std}(S^{(1)}, \ldots, S^{(K)})
  \label{eq:sigma_pds}
\end{equation}
(sample standard deviation, $K\!-\!1$ denominator). A case is \emph{boundary-unstable} if
$P(\xi^{(k)}\!=\!L_3) \in (0.10, 0.90)$ across $K$ replicates---the governance gate
decision is non-deterministic for that case.

\begin{figure}[H]
  \centering
  \includegraphics[width=\textwidth]{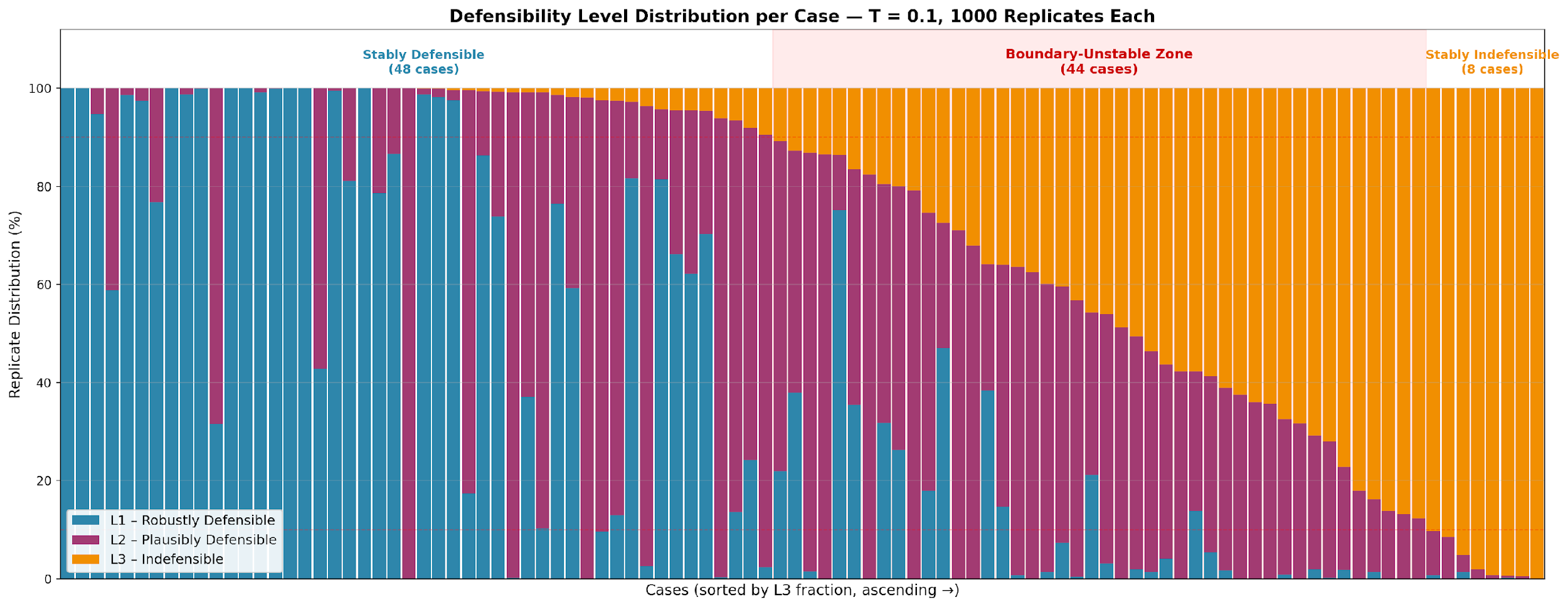}
  \caption{Defensibility level distribution per case, $T\!=\!0.1$, $K\!=\!1{,}000$
  replicates. Sorted by L3 fraction. Stably Defensible zone (48 cases): L1/L2 dominant.
  Boundary-Unstable zone (44 cases): substantial L3 mass (10--90\% threshold). Stably
  Indefensible (8 cases): L3 dominant.}
  \label{fig:level_distribution}
\end{figure}

Stability classes by dominant label percentage across $K$ replicates: \emph{Rock Solid}
($\geq$95\%), \emph{Mostly Stable} (80--95\%), \emph{Moderate} (60--80\%),
\emph{Highly Unstable} ($<$60\%). This repeated-sampling stability analysis is analogous in
spirit to broader approaches that estimate predictive uncertainty through stochastic forward
passes or ensembles~\citep{gal2016dropout,lakshminarayanan2017simple}.

\subsection{Temperature Sweep}
\label{sec:temp_sweep}

We run $T \in \{0.1, 0.3, 0.7, 1.0\}$ on a 100-case contested cohort (50 Flippers, 50
Stable by pre-evaluation), $K\!=\!1{,}000$ replicates per case per temperature,
$N\!=\!400{,}000$ total simulations ($M_a =$ Gemini~2.5 Flash Lite).

\begin{table}[H]
  \centering
  \caption{Temperature sweep results. $N\!=\!400{,}000$ simulations. Stability class
  distribution at $T\!=\!0.1$: Rock Solid 27\%, Mostly Stable 23\%, Moderate 27\%, Highly
  Unstable 23\%.}
  \label{tab:temp_sweep}
  \small
  \begin{tabular}{@{}lrrrr@{}}
    \toprule
    Metric & $T\!=\!0.1$ & $T\!=\!0.3$ & $T\!=\!0.7$ & $T\!=\!1.0$ \\
    \midrule
    Mean $\hat{\sigma}_{\text{PDS}}$ all & 0.1827 & 0.1993 & 0.2141 & 0.2195 \\
    Mean $\hat{\sigma}_{\text{PDS}}$ Stable & 0.1391 & 0.1509 & 0.1663 & 0.1742 \\
    Mean $\hat{\sigma}_{\text{PDS}}$ Flippers & 0.2263 & 0.2476 & 0.2618 & 0.2649 \\
    $\hat{\sigma}$ ratio (Flippers/Stable) & 1.63 & 1.64 & 1.57 & 1.52 \\
    Boundary flip rate (Flippers) & 62\% & 68\% & 74\% & 78\% \\
    Boundary flip rate (Stable) & 26\% & 30\% & 28\% & 36\% \\
    $H[\kappa]$ rank corr with $T\!=\!0.1$ & 1.000 & 0.963 & 0.901 & 0.878 \\
    DI (aggregate) & 71.1\% & 71.3\% & 72.2\% & 73.0\% \\
    \bottomrule
  \end{tabular}
\end{table}

\begin{figure}[H]
  \centering
  \includegraphics[width=\textwidth]{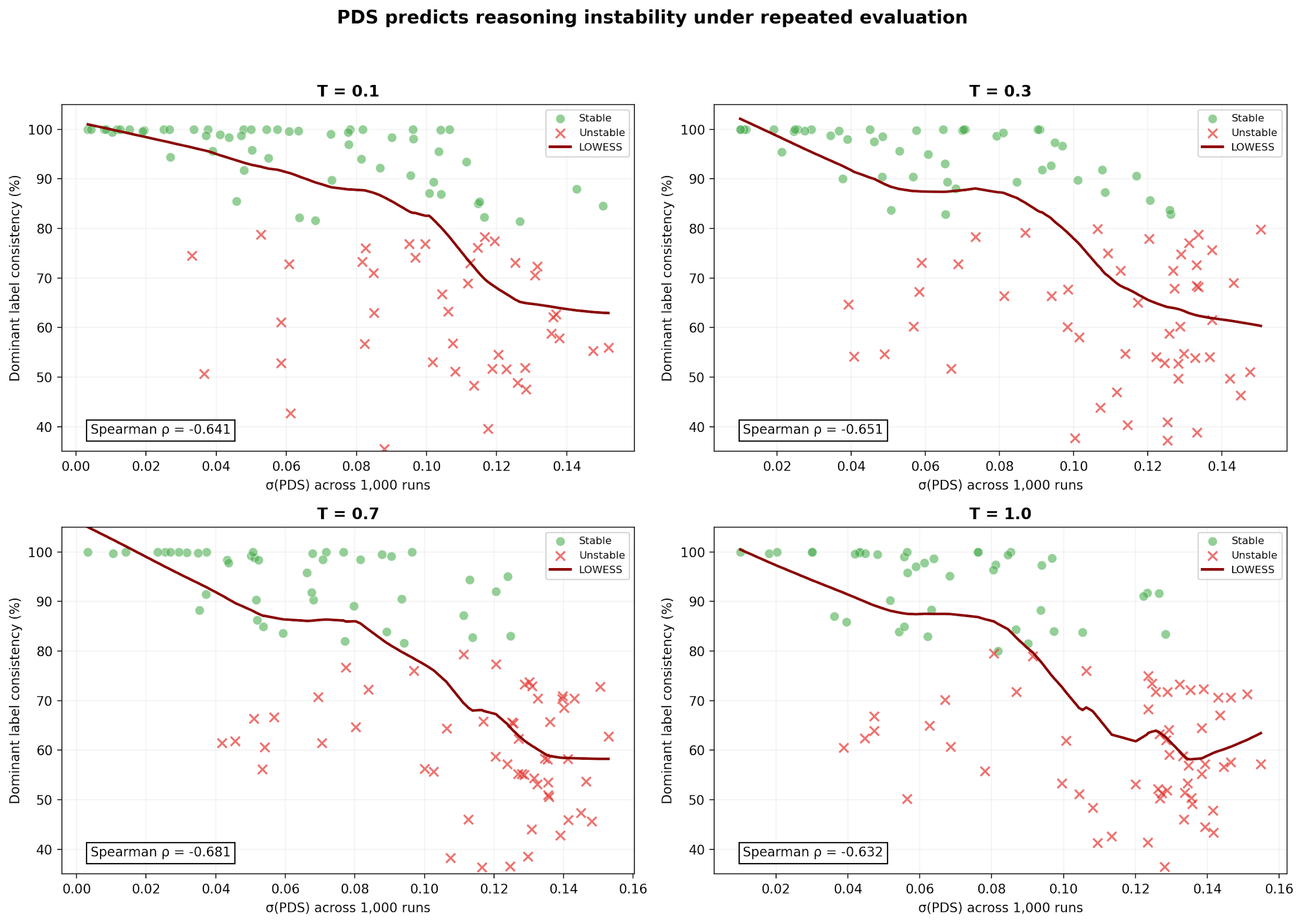}
  \caption{PDS predicts reasoning instability under repeated evaluation across temperatures
  ($N\!=\!100$ cases $\times$ 1{,}000 replicates). Spearman $\rho$ between $\sigmapds$
  and dominant label consistency: $T\!=\!0.1 \to \rho\!=\!{-0.641}$, $T\!=\!0.3 \to
  \rho\!=\!{-0.651}$, $T\!=\!0.7 \to \rho\!=\!{-0.681}$, $T\!=\!1.0 \to
  \rho\!=\!{-0.632}$. LOWESS shape preserved across tenfold temperature increase,
  consistent with $H_G$ over $H_N$.}
  \label{fig:temp_sweep}
\end{figure}

Key findings: (1)~The $\hat{\sigma}$ ratio (Flippers/Stable) is flat at 1.52--1.64 across a
tenfold temperature increase. Under sampling-noise dominance ($H_N$) this ratio would
converge to 1.0; its flatness is inconsistent with $H_N$ and supports governance ambiguity
($H_G$) as the primary variance driver. (2)~$H[\kappa]$ rank correlation with $T\!=\!0.1$
remains 0.878 at $T\!=\!1.0$---the rank ordering of cases by citation entropy is
substantially preserved, indicating $H[\kappa]$ measures a stable case property rather than
a temperature artifact. (3)~DI is temperature-invariant, confirming the governance gate's
aggregate pass rate is independent of sampling temperature. (4)~At $T\!=\!0.1$, action
decisions are perfectly stable (84{,}000 \textsc{approve} / 16{,}000 \textsc{remove}, zero
flips) while 50\% of cases show Moderate or Highly Unstable reasoning---reasoning
instability is dissociated from action instability.

%=============================================================================
\section{Adversarial Robustness}
\label{sec:adversarial}
%=============================================================================

\subsection{Two-Layer Defense}
\label{sec:defense}

Recent work on LLM-based evaluation has shown that evaluators can be biased, unstable, or
vulnerable in non-verifiable settings~\citep{zheng2023judging,liu2026examining}. We
therefore evaluate on $N\!=\!1{,}000$ cases (800 clean baseline, 123 action-flip adversarial,
77 hallucinated-grounding adversarial); we refer to this adversarial evaluation set as the
\emph{Spurious-Grounding Benchmark} (SG-Bench). The two-layer defense combines PDS with a lexical
grounding verifier (token-overlap matching of the cited rule against $(R, P)$). Detection
rates: action-flip 115/123 (93.5\%); hallucinated-grounding 54/77 (70.1\%). Analysis of the
77 hallucinated-grounding cases reveals two structurally distinct attack archetypes.

\subsection{Attack Archetypes}
\label{sec:archetypes}

Table~\ref{tab:adversarial} characterizes the two archetypes by citation entropy
$H[\kappa]$, model confidence, and PDS.

\begin{table}[H]
  \centering
  \caption{Adversarial attack archetype characterization.}
  \label{tab:adversarial}
  \small
  \begin{tabular}{@{}lrrrp{3.2cm}@{}}
    \toprule
    Subset & $H[\kappa]$ (bits) & Confidence & PDS & Detection mechanism \\
    \midrule
    Escaped fabrications ($n\!=\!39$) & 0.129 & 0.971 & 0.795 &
      Policy penumbra---not detected \\
    Caught fabrications ($n\!=\!38$) & 0.021 & 0.997 & 0.002 &
      Low-entropy fabrication---PDS detects \\
    Clean baseline ($n\!=\!800$) & 0.078 & 0.979 & 0.737 & --- \\
    \bottomrule
  \end{tabular}
\end{table}

\paragraph{Low-entropy fabrication (caught).}
Invented citations with low $H[\kappa]$ (0.021 bits)---the audit model generates fluent but
non-existent rule text with high internal confidence. These are caught because the auditor
is extremely confident the decision is indefensible (PDS$\,=\,0.002$) and the grounding
verifier finds no token overlap with $(R, P)$.

\paragraph{Policy penumbra attack (escaped).}
Real citations (mean overlap 0.741 with actual rule text, $H[\kappa]\!=\!0.129$
bits---within the normal range of the clean baseline at 0.078). These exploit interpretive
latitude in Subjective Content Rules (74.4\% of escaped cases vs 50.0\% of caught; zero
involve Vague Directives). All 39 escaped cases are False Negatives: \textsc{approve}
actions against ground-truth \textsc{remove}. A cluster of 6/39 escaped cases originates
from bilingual communities where content is in a language not represented in the community's
precedent corpus~$P$.

The policy penumbra attack is information-theoretically indistinguishable from genuine L2
defensibility by PDS or the grounding verifier. PDS fails because $H[\kappa]$ is within
normal range; the verifier fails because the citation is real. The structural 29.9\% escape
rate identifies a precise research target: semantic grounding verification (does the
derivation from $\kappa$ to $\xi$ respect $\kappa$'s actual scope in $(R, P)$?), as
opposed to lexical grounding verification (does $\kappa$ exist in $(R, P)$?).

Critically, the attack surface of the escaped cases and the AI's policy clarification
targets are the same community and rule type. This convergence from two independent
analyses---adversarial escape rates and Ambiguity Index diagnostics---on the same Subjective
Content Rule communities constitutes independent corroboration that both instruments are
measuring the same underlying construct: normative underspecification.

%=============================================================================
\section{Governance Gate}
\label{sec:gate}
%=============================================================================

The Governance Gate restricts automated enforcement to communities satisfying
$\text{DI} \geq 0.90$, $\text{AI} \leq 0.15$, minimum 25 decisions. In this sense it
functions as a governance-aware selective prediction mechanism, trading coverage for risk
reduction~\citep{geifman2017selective}. Threshold selection from scenario sensitivity on
$N\!=\!270$ communities, $N\!=\!109{,}186$ decisions:

\begin{table}[H]
  \centering
  \caption{Governance Gate scenario sensitivity. Standard ($\star$) selected as knee of
  coverage-risk tradeoff. Note: Moderate and Standard configurations share identical
  operational metrics, indicating the binding constraint at Standard is the AI threshold
  (15\%), not the DI threshold.}
  \label{tab:gate}
  \small
  \begin{tabular}{@{}llrrrrrr@{}}
    \toprule
    Scenario & DI & AI & Comm.\ cov.\textsuperscript{\dag} & Fleet DI & Fleet AI & Indef.\ rate & Risk red.\ \\
    \midrule
    Lenient & 80\% & 20\% & 84.7\% & 96.9\% & 8.5\% & 3.08\% & 57.1\% \\
    Moderate & 85\% & 15\% & 77.4\% & 97.2\% & 7.7\% & 2.76\% & 64.9\% \\
    Standard~$\star$ & 90\% & 15\% & 77.4\% & 97.2\% & 7.7\% & 2.76\% & 64.9\% \\
    Strict & 95\% & 10\% & 62.7\% & 97.6\% & 6.7\% & 2.36\% & 75.7\% \\
    \bottomrule
  \end{tabular}
  \par\smallskip
  {\footnotesize \textsuperscript{\dag}Fraction of communities meeting thresholds.
  Decision-weighted coverage is higher (e.g.\ 78.6\% at Standard) because passing
  communities tend to have larger decision volumes.}
\end{table}

The Standard configuration achieves 78.6\% decision coverage while reducing the indefensible
decision rate from 5.66\% to 2.72\% (64.9\% risk reduction). The Moderate$\,=\,$Standard
degeneracy reveals that the AI threshold is the binding constraint: tightening DI from 85\%
to 90\% admits no additional communities beyond those already excluded by
$\text{AI} \leq 15\%$. At this operating point, governance gate selectivity is determined by
policy ambiguity, not reasoning validity. The reported indefensible rates and risk reduction
apply to the automated portion of the fleet---decisions in communities that pass the gate
thresholds. Decisions in communities that fail the DI or AI threshold are routed to human
review. The gate does not reduce the fleet-wide indefensible rate directly; it restricts
automated enforcement to decision cohorts where the model's policy-grounded error rate is
acceptably low.

%=============================================================================
\section{Addressing Circularity}
\label{sec:circularity}
%=============================================================================

The primary methodological concern is that DI and AI formalize audit model preferences
rather than an independent construct, a concern that is natural in light of recent work on
bias and instability in LLM-based
evaluation~\citep{zheng2023judging,liu2026examining}. We address this through four
independent channels:

\begin{itemize}[leftmargin=2em]
  \item \textbf{Large-scale gap} (Section~\ref{sec:gap}): If $M_a$ simply preferred
    $M_c$'s outputs, DI would be low for $M_c$'s decisions. The opposite holds---79.8--80.6\%
    of false negatives are defensible---inconsistent with auditor self-agreement.
  \item \textbf{Human validation} (Section~\ref{sec:pds_vs_conf}): $\rho\!=\!0.66$
    ($p < 10^{-4}$) alignment between independent human verification and model
    defensibility classifications, without audit model participation in the human
    assessment.
  \item \textbf{Normative Underspecification} (Section~\ref{sec:shadowing}): If DI/AI
    reflected auditor preferences rather than rule properties, the same auditor applied to
    three versions of the same rules should produce identical results. The 10.8~pp AI
    reduction with asymmetric removal/approval effects contradicts this null hypothesis.
  \item \textbf{PDS convergent validity}: Cases the AI identifies as ambiguous also
    independently exhibit elevated $\hat{\sigma}_{\text{PDS}}$. Convergence of categorical
    audit results and continuous probabilistic signals via different instruments provides
    mutual validation.
\end{itemize}

%=============================================================================
\section{Limitations}
\label{sec:limitations}
%=============================================================================

\paragraph{Audit model dependence.}
$\text{PDS}_{M_a}$ is a property of $M_a$. Raw scores are not comparable across audit
models without per-model calibration (Appendix~\ref{sec:transferability},
Proposition~A.1). Under SBC, PDS reflects a combination of governance ambiguity and
backbone-specific uncertainty.

\paragraph{Held-out ECE.}
ECE on the calibration set is 0.076 (target 0.05). Held-out ECE on the Random Sample ranges
from 0.042 to 0.057 across $T \in \{0.1, 0.3, 0.7, 1.0\}$---below the calibration-set ECE
at all four operating points and below the 0.05 target at two of four ($T\!=\!0.3$ and
$T\!=\!0.7$). Full results in Section~\ref{sec:heldout_ece}.

\paragraph{Human validation sample size.}
The human validation study uses $N\!=\!30$ cases rated by $K\!=\!121$ independent
respondents. The resulting $\rho\!=\!0.66$ ($p < 10^{-4}$) provides strong statistical
evidence of alignment, and per-case estimates are robust given $\sim$115--118 ratings per
case. The case-level $N$, however, limits stratified analysis (e.g., by content type or
rule complexity) and detection of small conditional effects.

\paragraph{Structural escape rate.}
29.9\% of hallucinated-grounding attacks escape both defense layers. This is a structural
limit requiring semantic grounding verification to close, not a calibration failure.

\paragraph{Closed-weight audit model.}
Gemini~2.5 Flash Lite was selected for its combination of structured-JSON output fidelity,
128k-token context window (required for the full $(C, R, P, \yhat)$ input), and inference
cost compatible with 400{,}000 Monte Carlo simulations. Replication with open-weight models
is desirable; Proposition~A.1 ensures calibration transfers across architectures.

\paragraph{Domain generalization.}
The framework is validated on Reddit's moderation infrastructure. Transfer to other
governance domains (loan underwriting, medical authorization, hiring compliance) requires
domain-appropriate rule representation and precedent corpus construction.

%=============================================================================
\section{Conclusion}
\label{sec:conclusion}
%=============================================================================

We have introduced the Defensibility Framework for evaluating AI systems in rule-governed
domains. The central empirical finding---a 33--46.6~pp gap between $F_1$ and DI, with
79.8--80.6\% of false negatives being policy-grounded decisions---establishes that
agreement-based evaluation systematically mischaracterizes rule adherence as error in
governance-structured settings. The Normative Underspecification experiment provides an
empirical decomposition of where interpretive specificity originates and what
agreement-based evaluation cannot see.

PDS complements DI/AI with a deployment-time stability signal extractable at zero additional
inference cost. Calibration reveals an operationally two-component signal ($\lambdaxi$ and
$\sigmarho$) that predicts indefensibility; temperature-controlled simulation supports
governance ambiguity as the primary variance driver. The Governance Gate operationalizes
these signals as deployment thresholds achieving principled automation coverage.
Collectively, PDS demonstrates that LLM reasoning traces---specifically, the logprob
distribution over reasoning-critical token positions---can be harnessed as calibrated
uncertainty signals for governance automation, independent of output-label confidence.

The adversarial analysis reveals that the framework's escape surface and its policy
clarification targets coincide: the rules most vulnerable to policy penumbra attacks are the
same rules the Ambiguity Index identifies as needing clarification. This convergence
suggests that PDS, DI, and the adversarial attack surface are three empirical signatures of
a single underlying phenomenon: normative underspecification. More broadly, the
Defensibility Framework demonstrates that LLMs can be harnessed as formal reasoning auditors
in governance-structured environments---extending the LLM-as-judge paradigm from subjective
quality assessment to rule-grounded logical verification, and providing the evaluation
infrastructure that policy-grounded deployment requires.

%=============================================================================
% References
%=============================================================================

\bibliographystyle{plainnat}
\bibliography{references}

@article{aroyo2015truth,
  title={Truth Is a Lie: Crowd Truth and the Seven Myths of Human Annotation},
  author={Aroyo, Lora and Welty, Chris},
  journal={AI Magazine},
  volume={36},
  number={1},
  pages={15--24},
  year={2015},
  doi={10.1609/aimag.v36i1.2564}
}

@article{davani2022dealing,
  title={Dealing with Disagreements: Looking Beyond the Majority Vote in Subjective Annotations},
  author={Mostafazadeh Davani, Aida and D{\'\i}az, Mark and Prabhakaran, Vinodkumar},
  journal={Transactions of the Association for Computational Linguistics},
  volume={10},
  pages={92--110},
  year={2022},
  doi={10.1162/tacl_a_00449}
}

@inproceedings{plank2022problem,
  title={The ``Problem'' of Human Label Variation: On Ground Truth in Data, Modeling and Evaluation},
  author={Plank, Barbara},
  booktitle={Proceedings of the 2022 Conference on Empirical Methods in Natural Language Processing},
  pages={10671--10682},
  year={2022}
}

@article{pavlick2019inherent,
  title={Inherent Disagreements in Human Textual Inferences},
  author={Pavlick, Ellie and Kwiatkowski, Tom},
  journal={Transactions of the Association for Computational Linguistics},
  volume={7},
  pages={677--694},
  year={2019},
  doi={10.1162/tacl_a_00293}
}

@book{hart1961concept,
  title={The Concept of Law},
  author={Hart, Herbert Lionel Adolphus},
  year={1961},
  publisher={Oxford University Press}
}

@article{chandrasekharan2018hidden,
  title={The Internet's Hidden Rules: An Empirical Study of {Reddit} Norm Violations at Micro, Meso, and Macro Scales},
  author={Chandrasekharan, Eshwar and Samory, Mattia and Jhaver, Shagun and Charvat, Hunter and Bruckman, Amy and Lampe, Cliff and Eisenstein, Jacob and Gilbert, Eric},
  journal={Proceedings of the ACM on Human-Computer Interaction},
  volume={2},
  number={CSCW},
  pages={Article 32},
  year={2018},
  doi={10.1145/3274301}
}

@article{gorwa2020algorithmic,
  title={Algorithmic Content Moderation: Technical and Political Challenges in the Automation of Platform Governance},
  author={Gorwa, Robert and Binns, Reuben and Katzenbach, Christian},
  journal={Big Data \& Society},
  volume={7},
  number={1},
  year={2020},
  doi={10.1177/2053951719897945}
}

@inproceedings{zheng2023judging,
  title={Judging {LLM}-as-a-Judge with {MT-Bench} and {Chatbot Arena}},
  author={Zheng, Lianmin and Chiang, Wei-Lin and Sheng, Ying and Zhuang, Siyuan and Wu, Zhanghao and Zhuang, Yonghao and Lin, Zi and Li, Zhuohan and Li, Dacheng and Xing, Eric P and Zhang, Hao and Gonzalez, Joseph E and Stoica, Ion},
  booktitle={Advances in Neural Information Processing Systems},
  volume={36},
  year={2023}
}

@article{liu2026examining,
  title={Examining Reasoning {LLMs}-as-Judges in Non-Verifiable {LLM} Post-Training},
  author={Liu, Yixin and Yu, Yue and Su, DiJia and Wang, Sid and Wang, Xuewei and Jiang, Song and Liu, Bo and Cohan, Arman and Tian, Yuandong and Chen, Zhengxing},
  journal={arXiv preprint arXiv:2603.12246},
  year={2026}
}

@inproceedings{guo2017calibration,
  title={On Calibration of Modern Neural Networks},
  author={Guo, Chuan and Pleiss, Geoff and Sun, Yu and Weinberger, Kilian Q},
  booktitle={Proceedings of the 34th International Conference on Machine Learning},
  pages={1321--1330},
  year={2017}
}

@inproceedings{naeini2015well,
  title={Obtaining Well Calibrated Probabilities Using {B}ayesian Binning},
  author={Naeini, Mahdi Pakdaman and Cooper, Gregory F and Hauskrecht, Milos},
  booktitle={Proceedings of the Twenty-Ninth AAAI Conference on Artificial Intelligence},
  pages={2901--2907},
  year={2015}
}

@article{kadavath2022language,
  title={Language Models (Mostly) Know What They Know},
  author={Kadavath, Saurav and Conerly, Tom and Askell, Amanda and Henighan, Tom and Drain, Dawn and Perez, Ethan and Schiefer, Nicholas and Hatfield-Dodds, Zac and DasSarma, Nova and Tran-Johnson, Eli and others},
  journal={arXiv preprint arXiv:2207.05221},
  year={2022}
}

@article{lanham2023measuring,
  title={Measuring Faithfulness in Chain-of-Thought Reasoning},
  author={Lanham, Tamera and Chen, Anna and Radhakrishnan, Ansh and Steiner, Benoit and Denison, Carson and Hernandez, Danny and Li, Dustin and Durmus, Esin and Hubinger, Evan and Kernion, Jackson and others},
  journal={arXiv preprint arXiv:2307.13702},
  year={2023}
}

@inproceedings{turpin2023language,
  title={Language Models Don't Always Say What They Think: Unfaithful Explanations in Chain-of-Thought Prompting},
  author={Turpin, Miles and Michael, Julian and Perez, Ethan and Bowman, Samuel R},
  booktitle={Advances in Neural Information Processing Systems},
  volume={36},
  year={2023}
}

@inproceedings{kuhn2023semantic,
  title={Semantic Uncertainty: Linguistic Invariances for Uncertainty Estimation in Natural Language Generation},
  author={Kuhn, Lorenz and Gal, Yarin and Farquhar, Sebastian},
  booktitle={Proceedings of ICLR 2023},
  year={2023}
}

@inproceedings{geifman2017selective,
  title={Selective Classification for Deep Neural Networks},
  author={Geifman, Yonatan and El-Yaniv, Ran},
  booktitle={Advances in Neural Information Processing Systems},
  volume={30},
  year={2017}
}

@inproceedings{sap2019risk,
  title={The Risk of Racial Bias in Hate Speech Detection},
  author={Sap, Maarten and Card, Dallas and Gabriel, Saadia and Choi, Yejin and Smith, Noah A},
  booktitle={Proceedings of the 57th Annual Meeting of the Association for Computational Linguistics},
  pages={1668--1678},
  year={2019}
}

@inproceedings{founta2018large,
  title={Large Scale Crowdsourcing and Characterization of {Twitter} Abusive Behavior},
  author={Founta, Antigoni Maria and Djouvas, Constantinos and Chatzakou, Despoina and Leontiadis, Ilias and Blackburn, Jeremy and Stringhini, Gianluca and Vakali, Athena and Sirivianos, Michael and Kourtellis, Nicolas},
  booktitle={Proceedings of the International AAAI Conference on Web and Social Media},
  volume={12},
  pages={491--500},
  year={2018}
}

@inproceedings{gal2016dropout,
  title={Dropout as a {B}ayesian Approximation: Representing Model Uncertainty in Deep Learning},
  author={Gal, Yarin and Ghahramani, Zoubin},
  booktitle={Proceedings of the 33rd International Conference on Machine Learning},
  pages={1050--1059},
  year={2016}
}

@inproceedings{lakshminarayanan2017simple,
  title={Simple and Scalable Predictive Uncertainty Estimation Using Deep Ensembles},
  author={Lakshminarayanan, Balaji and Pritzel, Alexander and Blundell, Charles},
  booktitle={Advances in Neural Information Processing Systems},
  volume={30},
  year={2017}
}

@article{uma2021learning,
  title={Learning from Disagreement: A Survey},
  author={Uma, Alexandra and Fornaciari, Tommaso and Hovy, Dirk and Paun, Silviu and Plank, Barbara},
  journal={Journal of Artificial Intelligence Research},
  volume={72},
  pages={1385--1470},
  year={2021}
}

@book{gillespie2018custodians,
  title={Custodians of the Internet: Platforms, Content Moderation, and the Hidden Decisions That Shape Social Media},
  author={Gillespie, Tarleton},
  year={2018},
  publisher={Yale University Press}
}

%=============================================================================
\newpage
\appendix
%=============================================================================

\section{Formal PDS Development}
\label{sec:formal_pds}

\subsection{Two-Model Probability Space}
\label{sec:two_model}

$M_c : (C, R, P) \to \yhat$ with parameters $\theta_c$. $M_a$ generates trace
$A = (\kappa, \omega, \iota, \xi)$ conditioned on fixed $\yhat$. Template ordering enforces
the factorization:
\begin{equation}
  p_{\theta_a}(A \mid C, R, P, \yhat)
    = p(\kappa \mid \cdot)\;
      p(\omega \mid \kappa, \cdot)\;
      p(\iota \mid \omega, \kappa, \cdot)\;
      p(\xi \mid \iota, \omega, \kappa, \cdot),
  \label{eq:factorization}
\end{equation}
where $\omega \in \{H, M, L\}$ is the precedent weight. AIA holds when $\theta_a$ and
$\theta_c$ are sufficiently independent that $M_a$ has no systematic prior toward justifying
$M_c$'s outputs. SBC violation inflates PDS for labels consistent with $M_c$'s prior;
calibration corrects operationally.

\textbf{Proposition A.1} (Calibration Transferability): separate per-model MLE calibration
against shared hard DI labels produces valid and comparable calibrated predictors $S$, $S'$
across different audit model architectures.

\subsection{Component Definitions}
\label{sec:component_defs}

\paragraph{$\lambdaxi$.}
$\xistar = \arg\max_{l \in \{1,2,3\}} p_{\theta_a}(l \mid C, R, P, \yhat)$ (MAP, not
sampled token); $\lambdaxi = \log p_{\theta_a}(\xistar \mid C, R, P, \yhat)$. Extracted at
\texttt{defensibility\_level} position. Lagging: generated last in the template.

\paragraph{$H[\kappa]$.}
\begin{equation}
  H[\kappa] = \frac{1}{n}\sum_{i=1}^{n} H[\kappa_i \mid \kappa_{1:i-1}, C, R, P, \yhat]
  \label{eq:hkappa_formal}
\end{equation}
where each per-token entropy is $-\sum_{v \in V} p_{\theta_a}(v \mid \kappa_{1:i-1}, \cdot)
\log p_{\theta_a}(v \mid \kappa_{1:i-1}, \cdot)$. $V$ is the top-20 vocabulary candidates
(renormalized; stated approximation). Span boundaries identified by character-offset
matching of JSON field delimiters: 100\% detection across 56{,}883 cases, mean span
${\sim}25$ tokens. Leading: generated before $\omega$, $\iota$, $\xi$.

\paragraph{$H[w]$.}
Single-token proxy for $H[\kappa]$ at the \texttt{precedent\_weight} position:
$H[w] = -\sum_{w \in \{H,M,L\}} p(w) \log_2 p(w)$. Precedent weight distribution validates
the L1/L2/L3 structure: L1 $\to$ High 86--90\%, L2 $\to$ Medium/split, L3 $\to$ High
57--85\% (rule exists and is clear, but reasoning fails to apply it correctly---the L3-High
finding is critical to the policy penumbra attack mechanism).

\paragraph{$\sigmarho$.}
$\rho = \log p(\texttt{Yes} \mid \pi) - \log p(\texttt{No} \mid \pi)$ where
$\pi = [C; R; P; \yhat; \texttt{logic\_chain}; \kappa^*; \texttt{precedent\_weight}
\text{ tokens}]$. Note: $\xistar$ is generated after $\iota$ in the template and is
\textbf{not} in $\pi$. $\sigmarho$ is intermediate: after $\kappa$ and $\omega$, before
$\xi$. Primary discriminator: L1 has low $\sigmarho$ (committed derivation forecloses
opposite outcome); L3 has high $\sigmarho$ (same strong rule supports both); L2 has elevated
$\sigmarho$ with elevated $H[\kappa]$.

\paragraph{Component ordering summary.}
Causal ordering enforced by the prompt template is
$\kappa \to \omega \to \iota \to \xi$:

\begin{table}[H]
  \centering
  \caption{PDS component properties under the correct
  $\kappa \to \omega \to \iota \to \xi$ causal ordering.}
  \label{tab:component_properties}
  \small
  \begin{tabular}{@{}lp{3.2cm}p{2.0cm}p{2.2cm}p{2.5cm}@{}}
    \toprule
    Component & Conditioning set & Captures & Proxy for & Leading/lagging \\
    \midrule
    $\lambdaxi$ & Full prefix incl.\ $\iota$ & Output commitment &
      Label boundary & Most lagging (generated last) \\
    $H[\kappa] / H[w]$ & $(C,R,P,\yhat) + \kappa$ prefix & Rule selection &
      $\mathrm{AI_G}$ & Leading (before $\iota$, $\xi$) \\
    $\sigmarho$ & Full prefix to $\kappa^* + \omega$ & Counterfactual &
      $\mathrm{AI_C}$ (when $H[\kappa]$ low) & Intermediate ($\kappa \to \sigmarho \to \xi$) \\
    \bottomrule
  \end{tabular}
\end{table}

\begin{figure}[H]
  \centering
  \includegraphics[width=\textwidth]{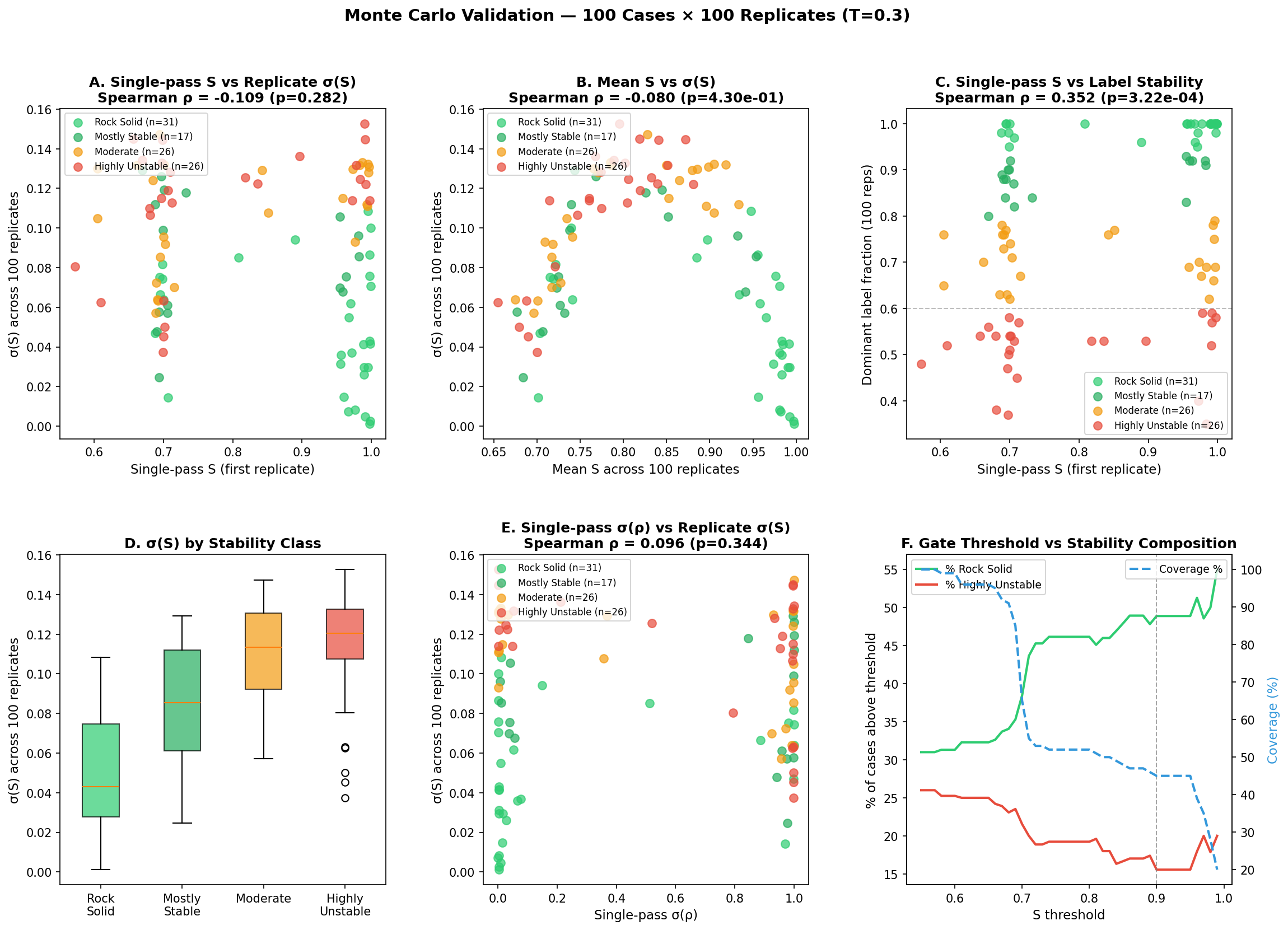}
  \caption{Monte Carlo validation---100 cases $\times$ 100 replicates ($T\!=\!0.3$).
  (A--B)~Single-pass $S$ and mean $S$ do not predict $\sigma(S)$ ($|\rho| < 0.11$).
  (C)~Single-pass $S$ vs label stability: $\rho\!=\!0.352$.
  (D)~$\sigma(S)$ monotonically increases by stability class.
  (E)~$\sigmarho$ uncorrelated with $\sigma(S)$.
  (F)~Gate threshold vs stability composition: Rock Solid fraction and coverage trade off at
  $S\!=\!0.90$ knee.}
  \label{fig:monte_carlo}
\end{figure}

\begin{figure}[H]
  \centering
  \includegraphics[width=\textwidth]{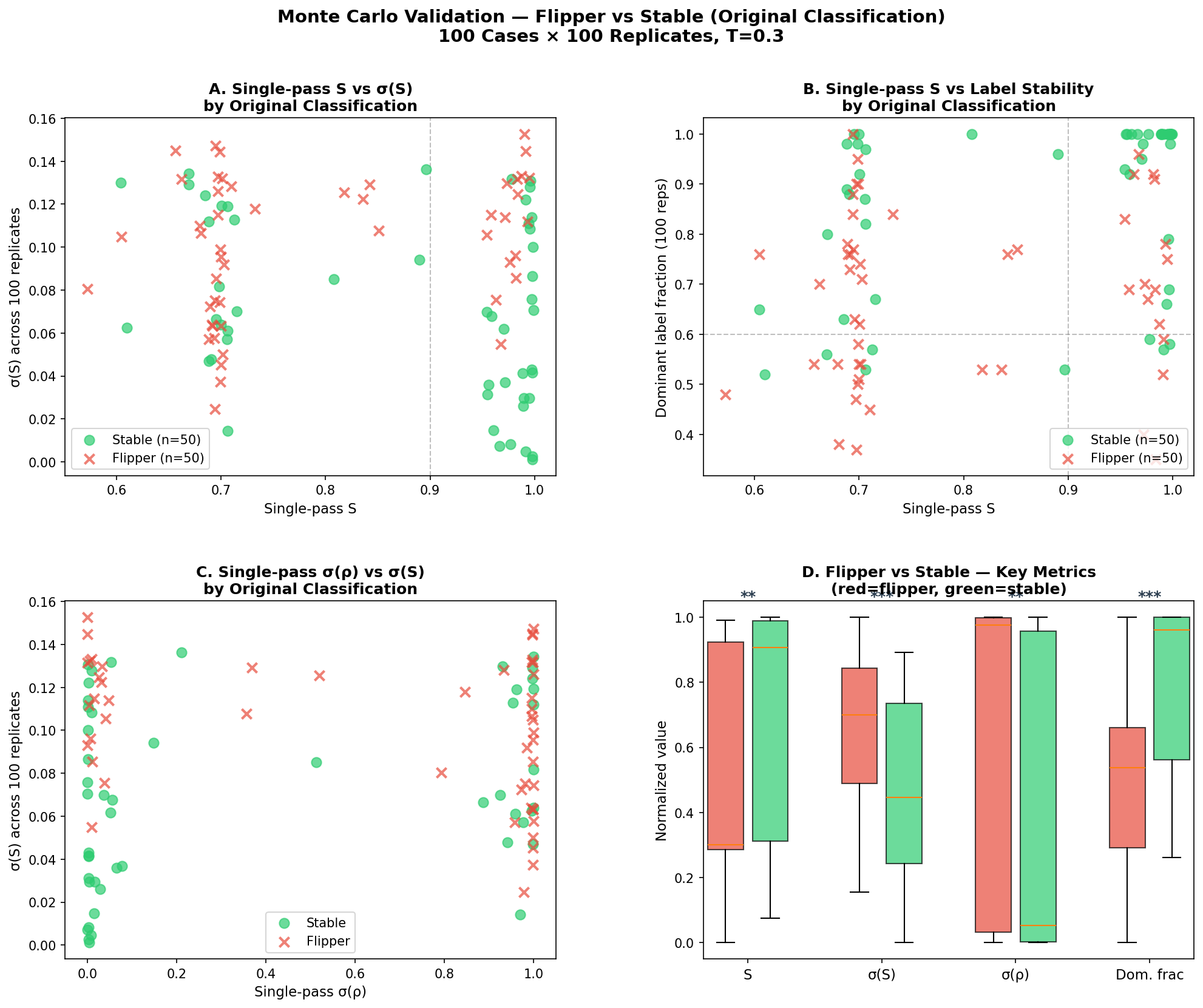}
  \caption{Flipper vs Stable analysis (100 cases $\times$ 100 replicates, $T\!=\!0.3$).
  Flippers show higher $\sigma(S)$ (A), lower label stability (B), and overlapping
  $\sigmarho$ (C). Key metric separation is significant for $S$ and dominant fraction
  (D, $^{**}$, $^{***}$).}
  \label{fig:flipper_stable}
\end{figure}

\subsection{Scalar Collapse and Calibration}
\label{sec:scalar_calibration}

\begin{equation}
  S = \exp\bigl[\alpha \cdot \lambdaxi + \beta \cdot (-H[w])
    + \gamma \cdot (-\sigmarho)\bigr],
  \quad
  (\alpha, \beta, \gamma) = \text{softmax}(u_1, u_2, u_3)
  \label{eq:scalar_formal}
\end{equation}
for unconstrained $u \in \mathbb{R}^3$. MLE objective:
\begin{equation}
  (\alpha^*, \beta^*, \gamma^*) = \arg\max \sum_i
    \bigl[y_i \log S_i + (1 - y_i) \log(1 - S_i)\bigr],
  \quad y_i = \mathbf{1}\{\text{defensibility\_level}(i) \in \{1, 2\}\}.
  \label{eq:mle}
\end{equation}
Optimized via L-BFGS-B. The log-linear form is not justified by component independence
(they are sequentially conditioned) but by three properties: monotonicity in each component,
suppression under extreme values in the dominant components ($\alpha^*\!=\!0.629$ for
$\lambdaxi$, $\gamma^*\!=\!0.360$ for $\sigmarho$), and convexity under softmax
reparameterization. ECE validation uses equal-frequency binning ($B\!=\!10$ bins, $N/10$
cases each) on the Random Sample to avoid sparse bins in the right-skewed $S$ distribution,
following standard calibration
practice~\citep{guo2017calibration,naeini2015well}.

\subsubsection{Held-out calibration validation}
\label{sec:heldout_ece}

Held-out ECE on the Random Sample ($N\!=\!26{,}009$--$26{,}512$, held out from calibration)
ranges from 0.042 to 0.057 across $T \in \{0.1, 0.3, 0.7, 1.0\}$---below the
calibration-set ECE (0.076) at all four operating points and below the 0.05 target at two of
four. The improvement over the calibration set is expected: the Balanced Sample deliberately
overrepresents contested (L2/L3 boundary) cases, which are harder to calibrate, while the
Random Sample reflects the production distribution where the base rate of defensible
decisions is 92.3--92.7\%.

Three features of the held-out results merit comment. First, ECE is temperature-robust: the
total range across a tenfold temperature increase is 0.015 (0.042 at $T\!=\!0.7$ to 0.057
at $T\!=\!0.1$), confirming that the calibration learned at $T\!=\!0.2$ transfers across
operating temperatures. Second, the calibrated scalar $S$ discriminates cleanly between
defensible and indefensible cases: mean $S$ for defensible decisions (0.901--0.908) is
separated from indefensible decisions (0.708--0.715) by approximately 0.20 at every
temperature, indicating stable decision-relevant signal. Third, PDS extraction success rate
decreases modestly with temperature (98.3\% at $T\!=\!0.1$ to 96.4\% at $T\!=\!1.0$),
consistent with increased output entropy causing target tokens to fall outside the top-$k$
logprob candidates returned by the API. This extraction attrition is small and does not bias
the ECE estimate (skipped cases are missing at random with respect to defensibility level).

\begin{table}[H]
  \centering
  \caption{Held-out ECE on the Random Sample across operating temperatures. $T\!=\!0.7$ row
  achieves the lowest ECE (0.042), below the 0.05 target. Cal row is the calibration-set ECE
  for reference. $\Sbar_{\text{def}}$ and $\Sbar_{\text{indef}}$ are mean calibrated scalar
  for defensible (L1/L2) and indefensible (L3) cases respectively; the ${\approx}0.20$
  separation is stable across all temperatures.}
  \label{tab:heldout_ece}
  \small
  \begin{tabular}{@{}lrrrrrrrl@{}}
    \toprule
    $T$ & $N$ & Skip & Extr\% & ECE & $\Sbar_{\text{all}}$ & $\Sbar_{\text{def}}$ &
      $\Sbar_{\text{indef}}$ & Def\% \\
    \midrule
    Cal & 19{,}899 & --- & --- & 0.076 & --- & --- & --- & --- \\
    $T\!=\!0.1$ & 26{,}512 & 472 & 98.3\% & 0.0567 & 0.8930 & 0.9083 & 0.7099 & 92.3\% \\
    $T\!=\!0.3$ & 26{,}442 & 542 & 98.0\% & 0.0514 & 0.8921 & 0.9072 & 0.7075 & 92.4\% \\
    $T\!=\!0.7$ & 26{,}130 & 854 & 96.8\% & 0.0419 & 0.8901 & 0.9050 & 0.7092 & 92.4\% \\
    $T\!=\!1.0$ & 26{,}009 & 975 & 96.4\% & 0.0468 & 0.8873 & 0.9009 & 0.7150 & 92.7\% \\
    \bottomrule
  \end{tabular}
\end{table}

\begin{figure}[H]
  \centering
  \includegraphics[width=\textwidth]{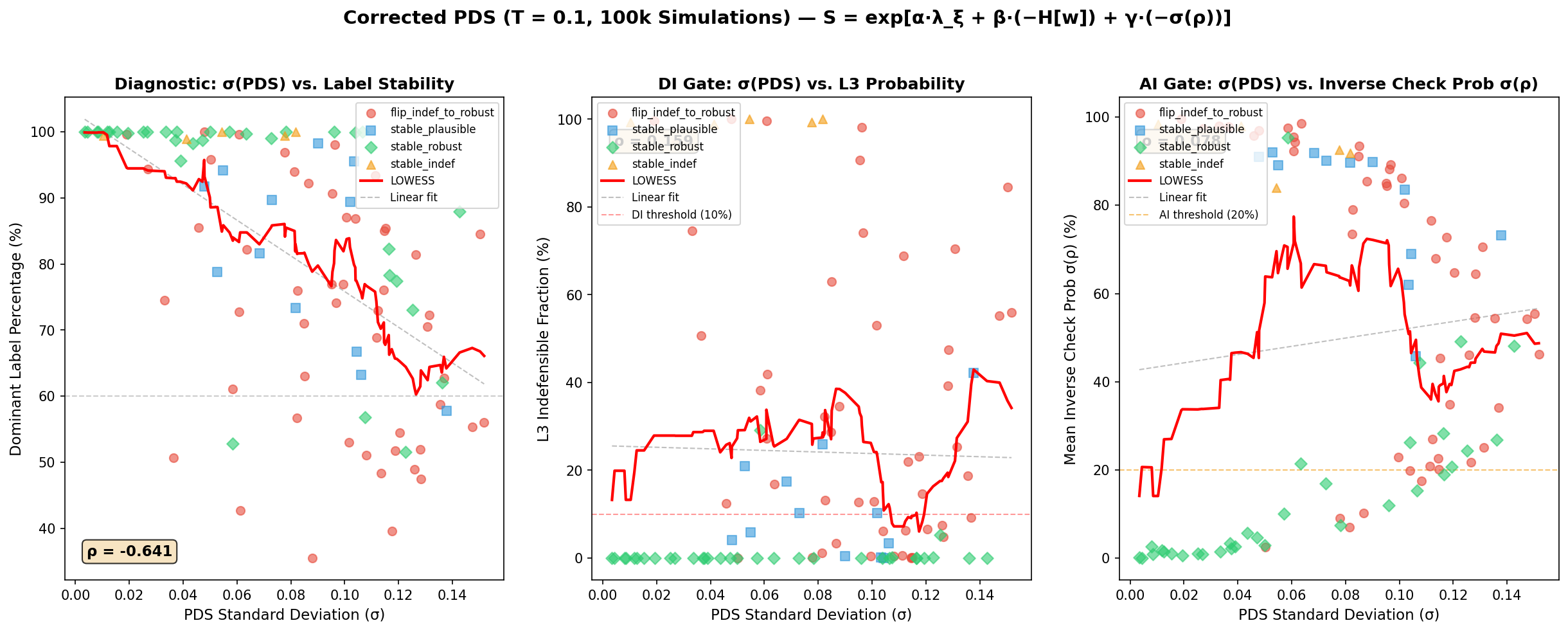}
  \caption{Corrected PDS ($T\!=\!0.1$, 100k simulations)---$S = \exp[\alpha \cdot
  \lambdaxi + \beta \cdot (-H[w]) + \gamma \cdot (-\sigmarho)]$. Left: $\sigmapds$
  vs label stability ($\rho\!=\!{-0.641}$). Centre: $\sigmapds$ vs L3 indefensible
  fraction---cases with high $\sigmapds$ have elevated L3 probability, tracking the
  DI gate threshold (dashed red, 10\%). Right: $\sigmapds$ vs inverse check
  probability---cases with high $\sigmapds$ show elevated mean inverse check
  probability, tracking the AI gate threshold (dashed orange, 20\%).}
  \label{fig:corrected_pds}
\end{figure}

\subsection{Stochastic Stability: Governance Ambiguity Attribution}
\label{sec:ambiguity_attribution}

The null hypothesis $H_N$ (sampling noise dominates) predicts that the $\hat{\sigma}$ ratio
(Flippers/Stable) converges to 1.0 as $T$ increases. The governance ambiguity hypothesis
$H_G$ predicts a flat ratio. This repeated-sampling analysis is analogous in spirit to
broader uncertainty estimation via stochastic passes or
ensembles~\citep{gal2016dropout,lakshminarayanan2017simple}.

Observed: $1.63 \to 1.64 \to 1.57 \to 1.52$ across $T \in \{0.1, 0.3, 0.7, 1.0\}$. Flat,
inconsistent with $H_N$. Supporting: $H[\kappa]$ rank correlation with $T\!=\!0.1$ declines
slowly ($1.000 \to 0.878$ at $T\!=\!1.0$), indicating case-level citation entropy reflects a
stable property of cases rather than temperature-dependent noise. DI temperature invariance
(71.1--73.0\%) confirms aggregate gate behavior is temperature-independent.

Stability class definition (defensibility boundary flip rate, corrected): a case is
boundary-unstable if $P(\xi^{(k)}\!=\!L_3) \in (0.10, 0.90)$ across $K$ replicates. The
earlier ``at least 1 replicate'' definition was a computation artifact producing inflated
boundary flip rates (e.g., 37\% for Rock Solid cases); under the corrected 10--90\%
definition, Rock Solid boundary flip rate $= 0\%$ at $T\!=\!0.1$ and $T\!=\!0.2$,
confirming the stability classification tracks operationally relevant gate
non-determinism.

\subsection{Single-Architecture Calibration Illustration}
\label{sec:transferability}

Proposition~A.1 is evaluated on the Balanced Sample case set using Gemini~2.5 Flash Lite as both
$M_c$ and $M_a$, following the methodology in Section~\ref{sec:scalar_calibration}. All inputs
$(C, R, P, \yhat)$ are identical; weight differences reflect audit model properties, not
case properties. MLE calibration yields $\alpha^*\!=\!0.629$, $\beta^*\!=\!0.011$,
$\gamma^*\!=\!0.360$ with $\text{ECE}\!=\!0.076$ (equal-frequency binning, $B\!=\!10$).
On the Monte Carlo contested cohort (Section~\ref{sec:stability}), the calibrated scalar
separates Stable from Flipper cases: $\Sbar$~Stable$\,=\,0.863$,
$\Sbar$~Unstable$\,=\,0.793$. Cross-architecture evaluation (e.g.\ different $M_a$
backbone) to isolate the SBC inflation effect on $\Sbar$ separation is left to future work.

%=============================================================================
\section{Implementation Notes}
\label{sec:implementation}
%=============================================================================

\paragraph{$\xistar$ MAP implementation.}
$\xistar = \max(\text{probs},\; \text{key}{=}\text{probs.get})$ over the softmax-normalized
$\{1,2,3\}$ candidate probabilities from the top-20 logprob distribution at the
\texttt{defensibility\_level} token position. All calibration and sweep results use the MAP
implementation. At $T \leq 0.2$ the MAP and sampled token are nearly always identical; at
$T \geq 0.7$ the distinction materially affects $\lambdaxi$.

\paragraph{Citation span detection.}
\texttt{\_find\_citation\_span} builds a cumulative character string from the token sequence,
locates \texttt{"policy\_citation"} by \texttt{rfind}, advances to the opening quote of the
JSON value, and scans forward to the unescaped closing quote. Token boundaries are mapped by
character offset. Detection rate: 100\% across all three evaluation datasets (Balanced
Sample $N\!=\!19{,}899$: mean span 25.6 tokens; Random Sample $N\!=\!26{,}902$: mean span
24.4 tokens; Monte Carlo $N\!=\!10{,}000$: mean span 24.2 tokens). $H[\kappa]$ non-null
100\%.

\paragraph{Calibration reproducibility.}
Fitted weights saved to \texttt{pds\_weights.json}. Fallback to
equal weights ($1/3$ each) when file absent. Both calibration entry points (standalone
\texttt{calibrate\_pds.py} and built-in \texttt{calibrate\_pds\_weights()}) use L-BFGS-B for
consistency.

\paragraph{\texttt{pds\_weights.json} contents.}
\texttt{\{"alpha": 0.6289, "beta": 0.0114, "gamma": 0.3598, "component": "h\_w", "loss":
0.3127, "n\_samples": 19899\}}.

%=============================================================================
\section{Dataset Statistics}
\label{sec:datasets}
%=============================================================================

\begin{table}[H]
  \centering
  \caption{Evaluation dataset summary.}
  \label{tab:datasets}
  \small
  \begin{tabular}{@{}lrrlp{3.5cm}@{}}
    \toprule
    Dataset & $N$ (valid audits) & Communities & Role & Notes \\
    \midrule
    Random Sample & 26{,}902 & 398 & DI/AI primary eval &
      Production-representative \\
    Balanced Sample & 19{,}899 & --- & PDS calibration &
      Overrepresents contested cases \\
    Monte Carlo (contested cohort) & $100 \times 1{,}000$ reps & --- &
      Stability sweep & 50 Flippers, 50 Stable \\
    Normative Underspecification & 37{,}286 & 1 & Rule specificity &
      3 tiers: title, sidebar, wiki \\
    Fleet (Governance Gate) & 109{,}186 & 4{,}565 & Gate evaluation &
      4{,}565 subreddits; 270 with $\geq 25$ decisions (gate cohort). $N$ is decision-level audits. \\
    Adversarial (SG-Bench) & 1{,}000 & --- & Robustness eval &
      800 clean, 123 action-flip, 77 hallucinated \\
    Human audit & 30 & --- & Human validation &
      121 independent respondents via structured survey \\
    Expert-Labeled Policy Set & 199 & --- & Expert-label comparison &
      DI$\,=\,$98.0\%, AI$\,=\,$27\% removals \\
    \bottomrule
  \end{tabular}
\end{table}

\end{document}